\documentclass[sigconf, nonacm]{acmart}

\copyrightyear{2023}
\acmYear{2023}
\setcopyright{cc}
\acmConference[CCS '23]{Proceedings of the 2023 ACM SIGSAC Conference on
Computer and Communications Security}{November 26--30, 2023}{Copenhagen,
Denmark}
\acmBooktitle{Proceedings of the 2023 ACM SIGSAC Conference on Computer
and Communications Security (CCS '23), November 26--30, 2023, Copenhagen,
Denmark}\acmDOI{10.1145/3576915.3623165}
\acmISBN{979-8-4007-0050-7/23/11}

\usepackage{graphicx, array, booktabs, tabularx, caption, subcaption, lipsum, enumitem, mathtools}
\usepackage{xcolor}

\definecolor{blue}{rgb}{0,0,0}

\newcommand{\defeq}{\vcentcolon=}
\newcolumntype{M}[1]{>{\centering\arraybackslash}m{#1}}
\renewcommand{\paragraph}[1]{
\vskip 1mm
\noindent
{\bf #1.}\hspace{.5em}
}

\usepackage{algorithm}
\usepackage{algpseudocode}
\algrenewcommand\algorithmicrequire{\textbf{Input:}}
\algrenewcommand\algorithmicensure{\textbf{Output:}}
\usepackage{balance}
\newcommand{\E}{\mathrm{E}}
\newcommand{\Var}{\mathrm{Var}}

\renewcommand{\epsilon}{\varepsilon}

\makeatletter
\def\subsubsection{\@startsection{subsubsection}{3}{\z@}%
  {-.5\baselineskip \@plus -2\p@ \@minus -.2\p@}%
  {.25\baselineskip}%
  {\ACM@NRadjust{\@subsubsecfont}}}
\makeatother

\theoremstyle{plain}
\newtheorem{theorem}{Theorem}[section]

\newtheorem{corollary}[theorem]{Corollary}
\theoremstyle{definition}
\newtheorem{definition}[theorem]{Definition}

\theoremstyle{remark}
\newtheorem{remark}[theorem]{Remark}

\begin{document}

\title{\textsc{Blink}: Link Local Differential Privacy in Graph Neural Networks via Bayesian Estimation}
\titlenote{This is an extended version of our paper to appear in CCS '23.}

\author{Xiaochen Zhu}
\orcid{0009-0007-2456-7406}
\affiliation{
  \institution{National University of Singapore}
  \city{}
  \country{}
}
\email{xczhu@nus.edu.sg}

\author{Vincent Y. F. Tan}
\orcid{0000-0002-5008-4527}
\affiliation{
  \institution{National University of Singapore}
  \city{}
  \country{}
}
\email{vtan@nus.edu.sg}

\author{Xiaokui Xiao}
\orcid{0000-0003-0914-4580}
\affiliation{
  \institution{National University of Singapore}
  \city{}
  \country{}
}
\email{xkxiao@nus.edu.sg}

\begin{abstract}
    Graph neural networks (GNNs) have gained an increasing amount of popularity due to their superior capability in learning node embeddings for various graph inference tasks, but training them can raise privacy concerns.
    To address this, we propose using link local differential privacy over decentralized nodes, enabling collaboration with an untrusted server to train GNNs without revealing the existence of any link.
    Our approach spends the privacy budget separately on links and degrees of the graph for the server to better denoise the graph topology using Bayesian estimation, alleviating the negative impact of LDP on the accuracy of the trained GNNs. We bound the mean absolute error of the inferred link probabilities against the ground truth graph topology. We then propose two variants of our LDP mechanism complementing each other in different privacy settings, one of which estimates fewer links under lower privacy budgets to avoid false positive link estimates when the uncertainty is high, while the other utilizes more information and performs better given relatively higher privacy budgets. Furthermore, we propose a hybrid variant that combines both strategies and is able to perform better across different privacy budgets. Extensive experiments show that our approach outperforms existing methods in terms of accuracy under varying privacy budgets.
\end{abstract}



\maketitle

\section{Introduction}

Graph neural networks (GNNs) achieve state-of-the-art performance in many domains, such as graph mining \cite{li2019graph}, recommender systems \cite{ying2018graph} and bioinformatics \cite{fout2017protein}. %
However, training GNNs can raise privacy concerns as the graph data used for training, such as social networks, may contain sensitive data that must be kept confidential as required by laws \cite{voigt2017eu}. Thus, there have recently garnered significant attention on the security and privacy of GNNs from the research community \cite{wu2022linkteller,he2021stealing,sajadmanesh2021locally,kolluri2022lpgnet,sajadmanesh2022gap}. Research has shown that neural networks can unintentionally leak information about training data \cite{shokri2017membership}, and there have been recent demonstrations of link inference attacks in GNNs \cite{wu2022linkteller,he2021stealing}. %
Hence, it is of particular significance to design privacy-preserving GNN frameworks.

\begin{figure}[!t]
    \centering
    \includegraphics[clip, trim=33mm 148mm 79mm 36mm, width=0.85\linewidth]{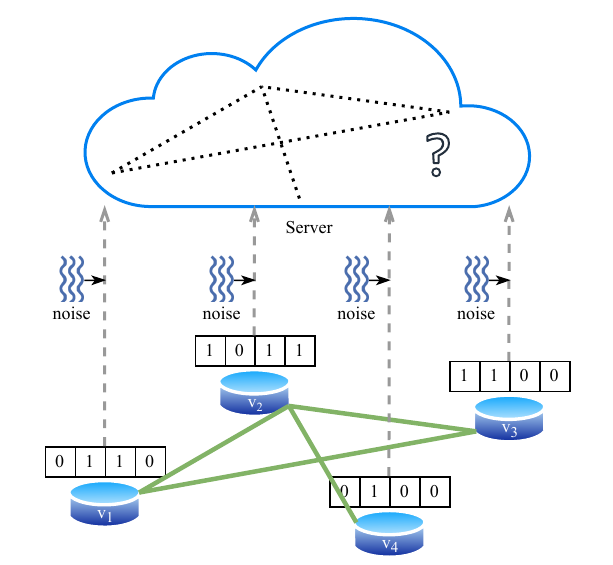}
    \caption{The problem of link local differential privacy over decentralized nodes. Each node first perturbs its adjacency list before sending to the server for privacy protection.}
    \Description{A figure to demonstrate the link LDP problem.}
    \label{fig:intro}
\end{figure}

Local differential privacy (LDP) \cite{cormode2018ldp, wang2019collecting,duchi2013local} is a rigorous privacy notion for collecting and analyzing sensitive data from decentralized data owners. Specifically, LDP ensures privacy by having each data owner perturb their data locally before sending it to the server, often through noise injection \cite{dwork2006calibrating}. The focus of our work is to design LDP mechanisms to protect graph topology (i.e., links) over decentralized nodes. In this setting, the server has access to the features and labels of all nodes, but not to any links among them. The server must infer the graph topology from the noisy adjacency lists transmitted by the nodes, as shown in Figure \ref{fig:intro}.

To illustrate the importance of link LDP in graph topology protection, consider a contact-tracing application installed on end devices for infectious disease control. The on-device application records interactions between other devices via Bluetooth, and the server trains a GNN to identify individuals at higher risk of virus exposure using the collected data. While local features, such as age and pre-existing conditions, can be voluntarily submitted by users and directly used by the server, this is not the case for contact history (i.e., links) due to the risk of revealing sensitive information such as users' whereabouts and interactions with others. Hence, it is crucial for end devices to perturb their links to achieve LDP before transmitting the information to the server for privacy protection.

\begin{figure}[!t]
  \centering
  \includegraphics[height=13pt]{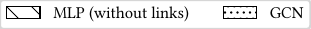}
  \vspace{5pt}
  \includegraphics{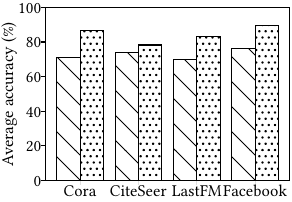}
  \vskip -0.3 cm 
  \caption{Test accuracy of GCN \cite{kipf2016semi} and MLP (GCN after removing links) on various graph datasets. Significant performance degeneration caused by removing links indicates the importance of graph topology in GNN training.}
  \label{fig:motivation}
\end{figure}

Our focus on link local privacy is driven by the following considerations. To start with, links represent the relationships between nodes, which data owners are often unwilling to disclose. Moreover, the issue of link LDP in GNNs over decentralized nodes as clients has yet to be sufficiently addressed in the literature, and there is currently a lack of effective mechanisms to balance privacy and utility. \cite{sajadmanesh2021locally} first propose locally differentially private GNNs, but only providing protection for node features and labels, while assuming the server has full access to the graph topology. Current differential privacy techniques for protecting graph topology while training GNNs, such as those described in \cite{lin2022towards,wu2022linkteller,hidano2022degree}, are limited by poor performance and are often outperformed by MLPs that are trained without link information at all (which naturally provides full link privacy). This issue with \cite{wu2022linkteller} has been investigated in \cite{kolluri2022lpgnet}, and we also demonstrate similar behaviors of other baselines in this paper.  On a separate line of research, there have been recent works on privacy-preserving graph synthesis and analysis with link local privacy guarantees \cite{qin2017generating, yelfgdpr2022, imola2021locally}. However, although some of these works do provide valid mechanisms to train GNNs with link LDP protections, these mechanisms are usually designed to estimate aggregate statistics of the graph, such as subgraph counts \cite{imola2021locally}, graph modularities and clustering coefficients \cite{yelfgdpr2022,qin2017generating}, which are not useful for training GNNs. Hence, these works are not directly applicable to our setting, and we will later show in this paper that they perform poorly in terms of GNN test accuracy. As such, there is a clear need for novel approaches to alleviate the performance loss of GNNs caused by enforcing privacy guarantees and to achieve link privacy with acceptable utility.

\begin{figure*}[!t]
    \centering
    \includegraphics[clip, trim=0mm 10mm 0mm 0mm, width=0.85\linewidth]{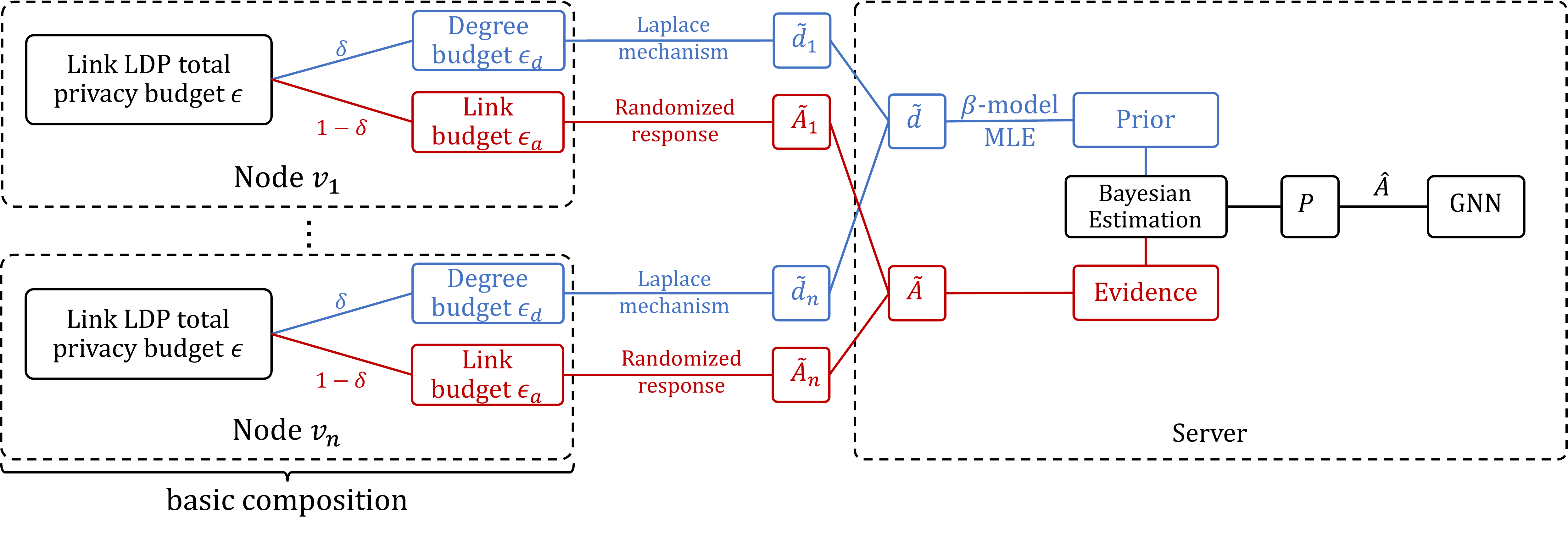}
    \vspace{-3mm}
    \caption{Structure of the proposed \textsc{Blink} framework.}
    \Description{A figure showing the structure of \textsc{Blink}. Each node separately injects noise to degree and adjacency list.}
    \label{fig:struct}
  \end{figure*}

\paragraph{Challenges} First, local DP is a stronger notion than central DP (CDP) where the magnitude of noise increases with the number of nodes. This creates an issue in real-world graph datasets where the number of vertices is typically large.
Moreover, as shown in Figure~\ref{fig:motivation}, removing links leads to a significant drop in GNN performance, indicating that graph topology is crucial in training effective graph neural networks. This is because GNN training is very sensitive to link alterations as every single wrong link will lead to the aggregation of information of neighboring nodes which should have been irrelevant. When the server adopts local differential privacy, it only has access to graph topology that is perturbed to be noisy for privacy protection, thus making it very challenging to train any effective GNNs on it.
Additionally, conventional LDP mechanisms such as randomized response \cite{warner1965rr} flip too many bits in the adjacency matrix and renders the noisy graph too dense, thus making it difficult to train any useful GNNs.
To conclude, it is challenging to alleviate the negative effects of local differential privacy on GNN performance. %

\paragraph{Contribution} In this paper, we propose \textsc{Blink} (\textbf{B}ayesian estimation for \textbf{link} local privacy), a principled mechanism for link local differential privacy in GNNs. Our approach involves separately and independently injecting noise into each node's adjacency list and degree, which guarantees LDP due to the basic composition theorem of differential privacy \cite{dwork2014algorithmic}. On the server side, our proposed mechanism uses Bayesian estimation to denoise the received noisy information in order to alleviate the negative effects on GNN performance of local differential privacy.

Receiving the noisy adjacency lists and degrees, the server first uses maximum likelihood estimation (MLE) in the $\beta$-model \cite{chatterjee2011random} to estimate the existence probability of each link based solely on the collected noisy degree sequence.
Then, the server uses the estimated link probability as a prior and the noisy adjacency lists as evidence to evaluate posterior link probabilities where both pieces of information are taken into consideration. We theoretically explain the rationale behind our mechanism and provide an upper bound of the expected absolute error of the estimated link probabilites against the ground truth adjacency matrix.
Finally, the posterior link probabilities are used to construct the denoised graph, and we propose three variants of such a construction---hard thresholding, soft thresholding, and a hybrid approach. Hard thresholding ignores links with a small posterior probability; it performs better when privacy budget is low and uncertainty is high because the lost noisy information would not significantly help with GNN training. The soft variant keeps all the inferred information and uses the posterior link probabilities as edge weights in the GNN model, and performs better than the hard variant when privacy budget is relatively higher thanks to the extra information. The hybrid approach combines both hard and soft variants and performs well for a wide range of privacy budgets. Extensive experiments demonstrate that all three variants of \textsc{Blink} outperform existing baseline mechanisms in terms of the test accuracy of trained GNNs. The hard and soft variants complement each other at different privacy budgets and the hybrid variant is able to consistently perform well across varying privacy budgets. %

\paragraph{Paper organization} The rest of this paper is organized as follows. Section \ref{sec:preliminaries} introduces preliminaries for GNNs and LDP and Section \ref{sec:statement} formally formulates our problem statement. We describe our proposed solution, \textsc{Blink}, in Section \ref{sec:approach} and explain its rationale and properties theoretically. We report and discuss extensive experimental results with all \textsc{Blink} variants and other existing methods in Section \ref{sec:experiments}. In Section \ref{sec:related}, we conduct a literature review on related topics and give brief introduction to relevant prior work. At last, Section \ref{sec:conclusion} concludes our work and discusses possible future research directions. The appendix includes complete proofs and experimental details.

\section{Preliminaries}\label{sec:preliminaries}

\subsection{Graph neural networks}

We consider the problem of semi-supervised node classification \cite{hamilton2017graphsage,kipf2016semi,velivckovic2017graph} on a simple undirected graph $G=(V,A,X,Y)$. Vertex set $V=\{v_i:i\in\{1,2,\ldots,n\}\}$ is the set of all $n$ nodes, consisting of labelled and unlabeled nodes. Let $V_L, V_U$ be the sets of labelled and unlabeled nodes, respectively, then $V_L\cap V_U=\emptyset$ and $V_L\cup V_U=V$. The adjacency matrix $A\in \{0,1\}^{n\times n}$ represents all the links in the graph, where $A_{i,j}=1$ if and only if a link exists between $v_i$ and $v_j$. The feature matrix of the graph is $X\in\mathbb{R}^{n\times d}$, where $d$ is the number of features on each node and for each $i$, row vector $X_i$ is the feature of node $v_i$. Finally, $Y\in\{0,1\}^{n\times c}$ is the label matrix where $c$ is the number of classes. In the semi-supervised setting, if vertex $v_i\in V_L$, then its label vector $Y_i$ is a one-hot vector, i.e. $Y_i\cdot\vec{1}$, where $\vec{1}$ is an all-ones vector of compatible dimension. Otherwise, when the vertex is unlabeled, i.e., $v_i\in V_U$, its label vector $Y_i$ is the zero vector $\vec{0}$.

A GNN learns high-dimensional representations of all nodes in the graph by aggregating node embeddings of neighbor nodes and mapping the aggregated embedding through parameterized non-linear transformation. More formally, let $x_i^{(k-1)}$ denote the node embedding of $v_i$ in $(k-1)$-th layer. The GNN learns the node embedding of $v_i$ in the $k$-th layer by\small
\begin{equation}\label{eq:gnn}
  x_i^{(k)}=\gamma^{(k)}(\text{Aggregate}(\{x_j^{(k-1)}:v_j\in\mathcal{N}(v_i)\}))
\end{equation}\normalsize
where $\mathcal{N}(v_i)$ is the set of neighboring nodes of $v_i$, $\text{Aggregate}(\cdot)$ is a differentiable, permutation invariant aggregation function such as sum or mean, and $\gamma(\cdot)$ is a differentiable transformation such as multi-layer perceptrons (MLPs). Note that the neighbor set $\mathcal{N}(v_i)$ may contain the node $v_i$ itself, depending on the GNN architecture. When initialized, all node embeddings are the node feature, i.e., $x_i^{(0)}=X_i$ for each $v_i\in V$. At the last layer, the model outputs vectors of $c$ dimension followed by a softmax layer to be compared against the ground truth so that the parameters in $\gamma$ can be updated via back-propagation to minimize a pre-defined loss function such as cross-entropy loss.

\subsection{Local differential privacy}

Differential privacy (DP) is the state-of-the-art mathematical framework to quantify and reduce information disclosure about individuals \cite{dwork2008dp,dwork2014algorithmic,yang2012dp}. DP bounds the influence of any individual tuple in the database to guarantee that one cannot infer the membership of any tuple from the released data, in a probabilistic sense. Usually, this is achieved by injecting noise to the data samples or the algorithm itself \cite{xiao2011differential, dwork2014algorithmic, dwork2006calibrating}. Mathematically, the most commonly used DP notion, $\epsilon$-differential privacy, is defined as follows.

\begin{definition}[$\epsilon$-DP]\label{def:dp}

  Let $\mathcal{D}$ be the space of all possible databases and $\mathcal{O}$ be the output space. A randomized algorithm $\mathcal{A}:\mathcal{D}\to\mathcal{O}$ is said to be $\epsilon$-differentially private if for any two databases $D,D'\in\mathcal{D}$ that only differ in exactly one record, and for any possible output $O\in\mathcal{O}$, we have\small
  \begin{equation}
    \frac{\Pr[\mathcal{A}(D)=O]}{\Pr[\mathcal{A}(D')=O]}\leq\exp(\epsilon).
  \end{equation}\normalsize

\end{definition}

In a central DP (CDP) setting, a data curator (server) applies a randomized algorithm $\mathcal{A}$ on a given database $D$ known to the curator and $\epsilon$-central DP is achieved if this central algorithm $\mathcal{A}$ satisfies Definition \ref{def:dp}. In a local model of DP \cite{wang2019collecting}, on the other hand, the data curator is untrusted and can only collect individual data from each data owner without being given the complete central database. Therefore, to preserve privacy, each data owner must implement a randomized algorithm to privatize its own data before transmitting to the server, and $\epsilon$-local DP (LDP) is achieved if each of such local randomizers satisfies Definition \ref{def:dp}. It is trivial to see that LDP is a stronger privacy notion where the server is no longer trusted, and more noise needs to be injected in LDP to achieve the same privacy budget as CDP.

\section{Problem Statement}\label{sec:statement}

We aim to protect the graph topology over decentralized nodes from an untrusted server. In our setting, each node stores information about itself and nothing of other nodes other than the existence of links, i.e., $v_i$ locally stores its feature vector $X_i$, its adjacency list $A_i$ and its label vector $Y_i$, and nothing else. Additionally, we assume that a server $S$ has access to $V$, $X$ and $Y$, but not the adjacency matrix $A$, which is kept private by the nodes. Collaborating with the nodes, the server tries to train a GNN model on $G$ to correctly classify the unlabeled nodes in $V_U$, without revealing the existence of any links in $G$. More specifically, we aim to design a local randomizer $\mathcal{R}$ to privatize the adjacency lists $A_i$ such that $\mathcal{R}$ achieves $\epsilon$-link LDP as defined below.

\begin{definition}[$\epsilon$-link LDP]\label{def:lldp}
  Randomized algorithm $\mathcal{R}: \{0,1\}^n\to\mathcal{O}$ is said to be $\epsilon$-link differentially private if for any two adjacency lists $a,a'\in\{0,1\}^n$ that only differ by one bit, i.e. $\Vert a-a'\Vert_{1,1}=1$, and for any possible outcome $O\in\mathcal{O}$, we have\small
  \begin{equation}
    \frac{\Pr[\mathcal{R}(a)=O]}{\Pr[\mathcal{R}(a')=O]}\leq\exp(\epsilon).
  \end{equation}\normalsize
\end{definition}

\begin{remark}
  Note that two adjacency lists are said to be neighbors if they differ by exactly one bit. This is the same as adding or removing exactly one edge in the graph. Therefore, if a mechanism satisfies $\epsilon$-link LDP as defined in Definition \ref{def:lldp}, the influence of any single link on the released output is bounded and thus the link privacy is preserved.
\end{remark}

After sending the privatized adjacency lists $\mathcal{R}(A_1),\ldots,\mathcal{R}(A_n)$ to the server, we also aim to design a server-side algorithm to denoise the received data which yields an estimated adjacency matrix $\hat{A}$. Finally, the server uses $(V,X,Y,\hat{A})$ to train a GNN to perform node classification as described in Equation (\ref{eq:gnn}). Additionally, note that although we assume that the server has access to $V,X,Y$, but it can be seen in Section \ref{sec:approach} that our proposed method does not involve the server utilizing node features or labels to denoise the graph topology. Hence, our method is compatible with existing LDP mechanisms that provide protections for node features and labels, such as LPGNN \cite{sajadmanesh2021locally}, and can serve as a convenient add-on to provide full local differential privacy on $X,Y,A$.

\section{Our approach}\label{sec:approach}

To train GNNs with link local differential privacy over decentralized nodes, we propose \textsc{Blink} (\textbf{B}ayesian estimation for \textbf{link} local privacy), a new framework to inject noise to the graph topology on the client side to preserve privacy and to denoise the server side to train better GNN models. The key idea is to independently inject noise to the adjacency matrix and degree sequence such that the degree of each node can be utilized by the server to better denoise the graph structure. More specifically, as shown in Figure~\ref{fig:struct}, the server uses the received noisy degree sequence as the prior and the noisy adjacency matrix as evidence to calculate posterior probabilities for each potential link. We now describe our method in more detail in the following subsections.

\subsection{Client-side noise injection}

As suggested by previous studies \cite{wu2022linkteller,lin2022towards,kolluri2022lpgnet}, simply randomly flipping the bits in adjacency lists will render the noisy graph too dense. Therefore, node degrees and graph density must be encoded in the private messages as well. Our main idea is to independently inject noise to the adjacency list and the degree of a node, and let the server estimate the ground truth adjacency matrix based on the gathered information. Based on this idea, we let the nodes send privatized adjacency lists and their degrees separately to the server, such that degree information can be preserved and utilized by the server to better denoise the graph topology. Specifically, for each node $v_i$, we spend the total privacy budget $\epsilon$ separately on adjacency list $A_i$ and its degree $d_i$, controlled by degree privacy parameter $\delta\in[0,1]$, such that we spend a privacy budget $\epsilon_d=\delta\epsilon$ on degree and the remaining $\epsilon_a=(1-\delta)\epsilon$ on adjacency list. This is possible because of the basic composition theorem of differential privacy \cite{dwork2014algorithmic}. For real-valued degree $d_i$, we use the widely-adopted Laplace mechanism \cite{dwork2006calibrating} to inject unbiased noise drawn from $\text{Laplace}(0, 1/\epsilon_d)$. And for bit sequence $A_i$, we use randomized response \cite{warner1965rr} to randomly flip each bit with probability $1/(1+\exp(\epsilon_a))$. This procedure is described in Algorithm \ref{alg:nodeldp}. According to basic composition and the privacy guarantee of Laplace mechanism and randomized response, we have the following theorem, stating that Algorithm \ref{alg:nodeldp} achieves $\epsilon$-link LDP. The detailed proof, together with the proofs of subsequent results, will be included in Appendix \ref{sec:proof}.

\begin{theorem}\label{thm:ldp}
  Algorithm \ref{alg:nodeldp} achieves $\epsilon$-link local differential privacy as defined in Definition \ref{def:lldp}.

\end{theorem}

\subsection{Server-side denoising}

After receiving the noisy adjacency lists $\tilde{A}_1,\tilde{A}_2,\ldots,\tilde A_n$ and degrees $\tilde d_1, \tilde d_2,\ldots,\tilde d_n$ from the nodes, the server first assembles them into a noisy adjacency matrix $\tilde A\in\{0,1\}^{n\times n}$ and noisy degree sequence $\tilde d\in\mathbb{R}^n$. The server then uses $\tilde d$ to estimate link probability to be used as prior, and uses $\tilde A$ as the evidence to calculate the posterior probability for each potential link to exist in the ground truth graph. At last, the server constructs graph estimations based on the posterior link probabilities and use the estimated graph to train GNNs. These steps are described in greater details as follows.

\subsubsection{Estimation of link probability given degree sequence}

{
\small
\begin{algorithm}[t]
  \caption{Node-side $\epsilon$-link LDP mechanism}\label{alg:nodeldp}
  \begin{algorithmic}[1]
    \Require $A_i\in\{0,1\}^{n}$ - adjacency list of $v_i$; $\epsilon$ - total privacy budget; $\delta$ - degree privacy parameter
    \Ensure $\left(\tilde{A}_i, \tilde{d}_i\right)$ - the private adjacency list $\tilde{A}_i\in\{0,1\}^{n}$ and the private degree $\tilde{d}_i\in\mathbb{R}$ of node $v_i$.
    \Function{LinkLDP}{$A_i,\epsilon,\delta$}:\Comment{run by node $v_i$}
    \State $\epsilon_d\gets \delta\epsilon$
    \State $\epsilon_a\gets (1-\delta)\epsilon$
    \For{$j\in\{1,2,\ldots,n\}$}
      \State $\tilde{A}_{ij}=
        \begin{cases}
          A_{ij}, & \text{with probability}\ {\exp(\epsilon_a)}/{(1+\exp(\epsilon_a))} \\
          1-A_{ij}, & \text{with probability}\ {1}/{(1+\exp(\epsilon_a))}
        \end{cases}$
    \EndFor\Comment{randomized response}
    \State $d_i\gets \sum_{j=1}^{n}A_{ij}$ \Comment{degree of node $v_i$}
    \State sample $l_i\sim\text{Laplace}(0,1/\epsilon_d)$
    \State $\tilde d_i\gets d_i+l_i$ \Comment{Laplace mechanism}
    \State \Return $\left(\tilde{A}_i,\tilde{d}_i\right)$
    \EndFunction
  \end{algorithmic}
\end{algorithm}
}

Given noisy degree sequence $\tilde d$, the server aims to estimate the probability of each link to exist, which is then used as prior probability in the next step. To estimate link probability, we adopt $\beta$-model, which is widely adopted in social network analysis \cite{chatterjee2011random,robins2007introduction,blitzstein2011sequential} and closely related to the well-known Bradley-Terry-Luce (BTL) model for ranking \cite{blt1952,blt1959,simons1999asymptotics,hunter2004mm}. Given a vector $\beta=(\beta_1,\beta_2,\ldots,\beta_n)\in\mathbb{R}^n$, the model assumes that a random undirected simple graph of $n$ vertices is drawn as follows: for each $1\leq i<j\leq n$, an edge between node $v_i$ and $v_j$ exists with probability \small\begin{equation}p_{ij}=\frac{\exp(\beta_i+\beta_j)}{1+\exp(\beta_i+\beta_j)}\end{equation}\normalsize independently of all other edges. Hence, the probability of observing the (true) degree sequence $d=(d_1,\ldots,d_n)$ from a random graph drawn according to the $\beta$-model is \small\begin{equation}
  L_d(\beta)=\frac{\exp(\sum_i \beta_i d_i)}{\prod_{i<j}(1+\exp(\beta_i+\beta_j))}.
\end{equation}\normalsize As a result, one can estimate the value of $\beta$ by maximizing the likelihood $L_d(\beta)$ of observing $d$. The maximum likelihood estimate (MLE) $\hat\beta$ of $\beta$ must satisfy the system of equations\small\begin{equation}\label{eqn:mleequasd}
  d_i=\sum_{j\neq i}\frac{\exp(\hat\beta_i+\hat\beta_j)}{1+\exp(\hat\beta_i+\hat\beta_j)},\qquad i=1,2,\ldots,n.
\end{equation}\normalsize
\citet{chatterjee2011random} show that with high probability, there exists a unique MLE solution $\hat\beta$ as long as the ground truth sequence $(\beta_i)$ is bounded from above and below, and the authors also provide an efficient algorithm for computing the MLE when it exists. Consider the following function $\phi_d:\mathbb{R}^n\to\mathbb{R}^n$ where \small\begin{equation}
  \phi_d(x)_i=\log(d_i)-\log\left(\sum_{j\neq i}\frac{1}{\exp(-x_j)+\exp(x_i)}\right).
\end{equation}\normalsize \citet{chatterjee2011random} prove that the MLE solution is a fixed point of the function $\phi$ and hence can be found iteratively using Algorithm~\ref{alg:mle}.

Therefore, if the degree sequence $d$ were to be released to and observed by the server, the server could then model the graph using the $\beta$-model and estimate link probabilities via MLE. However, actual degree sequence $d$ must be kept private to the server for the privacy guarantee. As per Algorithm \ref{alg:nodeldp}, $d$ is privatized through Laplace mechanism and only the noisy $\tilde d$ can be observed by the server. Although the server cannot directly maximize the likelihood $L_d(\beta)$ of observing $d$, the following theorem shows that the observable log-likelihood $\ell_{\tilde d}(\beta)=\log(L_{\tilde d}(\beta))$ is a lower bound of the unobservable $\ell_d(\beta)=\log(L_d(\beta))$ (up to a gap).

\begin{theorem}\label{thm:mle}
  For any $\beta\in\mathbb{R}^n$ that is bounded from above and below, let $M=\max_{1\leq i\leq n}|\beta_i|$. For any given constant $a$, with probability at least $1-1/a^2$, we have $\ell_{d}(\beta) \geq \ell_{\tilde d}(\beta)-a\sqrt nM/\epsilon_d^2$, where the probability is measured over the randomness of Laplace mechanism of $\tilde d$.
\end{theorem}

\small
\begin{algorithm}[t]
  \caption{MLE of link probability given degree sequence}\label{alg:mle}
  \begin{algorithmic}[1]
    \Require $d\in\mathbb{R}^n$ - degree sequence
    \Ensure $p\in[0,1]^{n\times n}$ - link probability matrix, where $p_{ij}$ is the estimated probability that an edge exists between $v_i$ and $v_j$
    \Function{MLELinkProbability}{$d$}:
    \State {initialize} $\beta\in \mathbb{R}^n$ as a zero vector
    \While{not converging}
      \State$\beta\gets\phi_d(\beta)$\Comment{MLE solution is a fixed point of function $\phi_d$ \cite{chatterjee2011random}}
    \EndWhile
    \For{$(i,j) \in \{1,2,\ldots,n\}^2 \land i\neq j$}
      \State $\displaystyle p_{ij}\gets\frac{\exp(\beta_i+\beta_j)}{1+\exp(\beta_i+\beta_j)}$ \Comment{$\beta$-model}
    \EndFor
    \State $p$.SetDiagonal$(0)$ \Comment{$\beta$-model does not consider self loops}
    \State\Return $p$
    \EndFunction
  \end{algorithmic}
\end{algorithm}\normalsize 

\begin{remark}
  Maximizing the observable but noisy log-likelihood $\ell_{\tilde{d}}(\beta)$ can maximize the unobservable target log-likelihood $\ell_d(\beta)$ to a certain extent (a gap of $\Theta(\sqrt nM/\epsilon_d^2)$). Similar to (\ref{eqn:mleequasd}), the solution $\hat\beta$ that maximizes log-likelihood $\ell_{\tilde{d}}(\beta)$ satisfies the following system of equations \small\begin{equation}\label{eqn:d'mlecondition}
    \tilde d_i=\sum_{j\neq i}\frac{\exp(\hat\beta_i+\hat\beta_j)}{1+\exp(\hat\beta_i+\hat\beta_j)},\qquad i=1,2,\ldots,n,
  \end{equation}\normalsize and it will be a fixed point of function $\phi_{\tilde d}$. However, Eq. (\ref{eqn:d'mlecondition}) has a solution only if $\tilde d_i\in(0,n-1)$ for all $i=1,2,\ldots,n$. Therefore, the server first clips the values of $\tilde d$ to $\tilde d^+\in[1,n-2]^n$ and then calls the function \textproc{MLELinkProbability}$(\tilde d^+)$ from Algorithm \ref{alg:mle} to find the link probabilities that maximize $\ell_{\tilde d^+}(\beta)$. This step is described in Lines 2 and 3 of Algorithm \ref{alg:be}.
\end{remark}

\subsubsection{Estimation of posterior link probabilities}\label{sec:post}

\small\begin{algorithm}[t]
  \caption{Server-side estimation of posterior link probabilities}\label{alg:be}
  \begin{algorithmic}[1]
    \Require $\tilde{A}\in\{0,1\}^{n\times n}, \tilde{d}\in\mathbb{R}^n$ - privatized adjacency matrix and degree sequence, where for $1\leq i\leq n$, $(\tilde{A}_i,\tilde d_i)=\textproc{LinkLDP}(A_i,\epsilon,\delta)$ executed by node $v_i$; $\epsilon$ - total privacy budget; $\delta$ - degree privacy parameter
    \Ensure $P\in[0,1]^{n\times n}$ - the posterior probabilities for each link to exist
    \Function{Denoise}{$\tilde{A},\tilde d,\epsilon,\delta$}:
    \State $\tilde{d}^+\gets \tilde d$.Clip$(\text{min}=1,\text{max}=n-2)$
    \State $p\gets \textproc{MLELinkProbability}(\tilde{d}^+)$\Comment{prior}
    \For{$(i,j)\in\{1,2,\ldots,n\}^2$}
      \State $\displaystyle q_{ij}\gets \Pr[(\tilde A_{ij},\tilde A_{ji})\vert(A_{ij},A_{ji})=(1,1)]$\Comment{evidence likelihood}
      \State $\displaystyle q'_{ij}\gets \Pr[(\tilde A_{ij},\tilde A_{ji})\vert(A_{ij},A_{ji})=(0,0)]$\Comment{evidence likelihood}
      \State $\displaystyle P_{ij}\gets \frac{q_{ij} p_{ij}}{q_{ij} p_{ij} + q'_{ij} (1-p_{ij})}$\Comment{Bayesian posterior probability}
    \EndFor
    \State \Return $P$
    \EndFunction
  \end{algorithmic}
\end{algorithm}\normalsize

The noisy degree sequence $\tilde d$ enables the server to estimate the link probabilities to be used as a prior, such that the server can use the received noisy adjacency matrix as evidence to evaluate posterior probability. For each potential link between $v_i,v_j\in V$, the server receives two bits $\tilde{A}_{ij}$ and $\tilde A_{ji}$ related to its existence. Because the privacy budget $\epsilon_a$ in RR (Algorithm \ref{alg:nodeldp}) is known to the server, the server can use the flip probability $p_\text{flip}=1/(1+\exp(\epsilon_a))$ to calculate the likelihood of observing the received bits $(\tilde{A}_{ij},\tilde A_{ji})$ conditioned on whether a link exists between $v_i$ and $v_j$ in the actual graph. More specifically, we have \small$$\begin{aligned}
  q_{ij}%
  &=\begin{cases}\label{eqn:qij}
    p_\text{flip}^2, &\quad(\tilde A_{ij}, \tilde A_{ji})=(0,0)\\
    p_\text{flip}(1-p_\text{flip}), &\quad(\tilde A_{ij}, \tilde A_{ji})=(0,1) \lor (\tilde A_{ij}, \tilde A_{ji})=(1,0)\\
    (1-p_\text{flip})^2, &\quad(\tilde A_{ij}, \tilde A_{ji})=(1,1)
  \end{cases};\\
  q'_{ij}%
  &=\begin{cases}\label{eqn:q'ij}
    (1-p_\text{flip})^2, &\quad(\tilde A_{ij}, \tilde A_{ji})=(0,0)\\
    p_\text{flip}(1-p_\text{flip}), &\quad(\tilde A_{ij}, \tilde A_{ji})=(0,1) \lor (\tilde A_{ij}, \tilde A_{ji})=(1,0)\\
    p_\text{flip}^2, &\quad(\tilde A_{ij}, \tilde A_{ji})=(1,1)
  \end{cases}.
\end{aligned}$$\normalsize
Here, $q_{ij}$ is the likelihood of observing $(\tilde A_{ij}, \tilde A_{ji})$ given the existence of the link $(v_i,v_j)$, and $q'_{ij}$ is the likelihood of observing $(\tilde A_{ij}, \tilde A_{ji})$ given the non-existence of the link $(v_i,v_j)$. Hence, together with the link probability (without taking evidence into consideration) $p_{ij}$ estimated solely from noisy degree sequence, one can apply Bayes rule to evaluate the posterior probability, i.e. for each $1\leq i\neq j\leq n$, \small\begin{align*}
  P_{ij}=\Pr[(A_{ij},A_{ji})=(1,1)|(\tilde A_{ij},\tilde A_{ji})]
  =\frac{q_{ij}p_{ij}}{q_{ij}p_{ij}+q'_{ij}(1-p_{ij})}.
\end{align*}\normalsize
For each $1\leq i\neq j\leq n$, $P_{ij}$ is the posterior probability that a link exists between $v_i$ and $v_j$ conditioned on the evidence $(\tilde A_{ij},\tilde A_{ji})$. We will show the accuracy of this estimation of graph topology in Section \ref{sec:theo} by bounding the mean absolute error between $P$ and ground truth $A$. %

\subsubsection{Graph estimation given posterior link probabilities}\label{sec:variants}
After obtaining $P$, we propose three different variants of \textsc{Blink} for the server to construct graph estimations used for GNN training.
\paragraph{\textsc{Blink-Hard}} The simplest and most straightforward approach is to only keep links whose posterior probability of its existence triumphs that of its absence, i.e. keep a link between $v_i$ and $v_j$ in the estimated graph $\hat A$ if and only if $P_{ij}>0.5$.

It is clear that hard-thresholding loses a lot of information contained in $P$ by simply rounding all entries to $0$ or $1$. However, when privacy budget is low and uncertainty is high, the information provided by the nodes are usually too noisy to be useful for GNN training, and may even corrupt the GNN model \cite{kolluri2022lpgnet}. Therefore, \textsc{Blink-Hard} is expected to perform better when privacy budget is low, while when privacy budget grows, it is likely to be outperformed by other variants of \textsc{Blink}.

\paragraph{\textsc{Blink-Soft}} Instead of hard-thresholding, the server can keep all the information in $P$ instead by using them as edge weights. In this way, the GNN formulation in (\ref{eq:gnn}) is modified as follows to adopt weighted aggregation:
\small\begin{equation}\label{eq:weighted_gnn}
  x_i^{(k)}=\gamma^{(k)}(\text{Aggregate}(\{P_{ij}\cdot x_j^{(k-1)}:v_j\in V\})),
\end{equation}\normalsize
where $\text{Aggregate}(\cdot)$ is a permutation invariant aggregation function such as sum or mean. Detailed modifications of specific GNN architectures will be included in Appendix \ref{sec:app-experiments}.

The soft variant utilizes extra information of $P$ compared to \textsc{Blink-Hard}, and hence, is expected to achieve better performance as long as the information is not too noisy to be useful. Therefore, we form a hypothesis that \textsc{Blink-Soft} and \textsc{Blink-Hard} complement each other and the former is preferred when privacy budget is relatively higher while the latter is preferred at lower privacy budgets.

\paragraph{\textsc{Blink-Hybrid}}
At last, we combine both the hard and soft variants such that the server can eliminate unhelpful noisy information while utilizing more confident information in $\Vert P\Vert_{1,1}$ via weighted aggregation. The server first takes the highest $\Vert P\Vert_{1,1}$ entries of $P$ and filters out the remaining by setting them as zeros. This is to only keep the top $\Vert P\Vert_{1,1}$ possible links in the graph as $\Vert P\Vert_{1,1}$ is an estimation of the graph density $\Vert A\Vert_{1,1}$ (suggested by Theorem \ref{thm:p_a_error} and Corollary \ref{col:mae}). This step is inspired by the idea of only keeping the top $|E|$ links from \textsc{DpGCN} \cite{wu2022linkteller}. Then, the server utilizes the remaining entries in $P$ by using them as edge weights and trains the GNN as in Equation (\ref{eq:weighted_gnn}). \textsc{Blink-Hybrid} is expected to incorporate the advantages of both \textsc{Blink-Hard} and \textsc{Blink-Soft} and perform well for all privacy levels.

\subsection{Theoretical analysis for utility}\label{sec:theo}

\begin{figure*}[hbt!]
  \centering
  \includegraphics{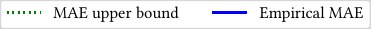}
  \vspace{5pt}

  \begin{subfigure}[b]{0.25\textwidth}
      \centering
      \includegraphics{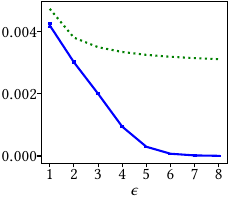}
      \vskip -0.2cm
      \caption{Cora}
      \label{fig:mae_cora}
  \end{subfigure}%
  \begin{subfigure}[b]{0.25\textwidth}
      \centering
      \includegraphics{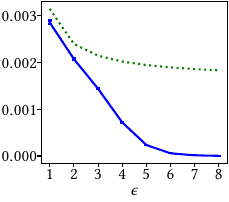}
      \vskip -0.2cm
      \caption{CiteSeer}
      \label{fig:mae_citeseer}
  \end{subfigure}%
  \begin{subfigure}[b]{0.25\textwidth}
      \centering
      \includegraphics{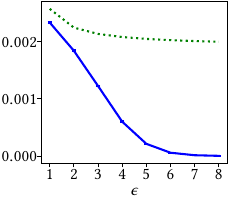}
      \vskip -0.2cm
      \caption{LastFM}
      \label{fig:mae_lastfm}
  \end{subfigure}%
  \begin{subfigure}[b]{0.25\textwidth}
    \centering
    \includegraphics{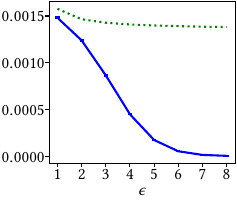}
    \vskip -0.2cm
    \caption{Facebook}
    \label{fig:mae_facebook}
\end{subfigure}
  \vskip -0.3cm
  \caption{Average mean absolute error (MAE) of the inferred link probabilities $P$ against ground truth $A$ ($\delta$ set to $0.1$).}
  \label{fig:mae}
\end{figure*}

We have described our proposed approach, namely, \textsc{Blink}, in detail in previous sections. While its privacy guarantee has been shown in Theorem \ref{thm:ldp}, we now theoretically demonstrate its utility guarantees.

\paragraph{Choice of utility metric} To quantify the utility of \textsc{Blink}, we first need to identify a metric to be bounded that is able to reflect the quality of the estimated graph, $P$. In the related literature, many metrics have been used to demonstrate the utility of differentially private mechanisms for graph analysis. For example, \citet{hidano2022degree} show that their estimated graph topology preserves the graph density; \citet{imola2021locally} bound the error in triangle count and $k$-star count; \citet{yelfgdpr2022} evaluate the error in any arbitrary aggregate graph statistic, as long as the aggregate statistic can be estimated without bias from both degree and neighbor lists, such as clustering coefficient and modularity. However, none of these metrics can reflect the quality of the estimated graph topology directly and represent the performance of the GNN trained on the estimated graph because they only involve aggregate and high-level graph statistics. In contrast, the performance of GNNs for node classification is very sensitive to link information from a microscopic or local perspective, as node information propagates along links and any perturbation of links will lead to aggregation of other nodes' information that should have been irrelevant, or missing the information of neighboring nodes. This is one of the reasons that many prior works involving privacy-preserving GNNs for node classification \cite{wu2022linkteller,lin2022towards,kolluri2022lpgnet} only provide empirical evidence of the utility of their approaches. Although there is no metric that can directly reflect the performance of the trained GNNs, the closer the estimated adjacency matrix $P$ is to the ground truth $A$, the better the GNNs trained on $P$ are expected to perform, and the closer the trained GNNs would perform compared to those trained with accurate graph topology. Therefore, we evaluate the utility of \textsc{Blink} as a statistical estimator of the ground truth adjacency matrix $A$ by bounding the expectation of the $\ell_1$-distance between $P$ and $A$, i.e. $\E[\Vert P-A\Vert_{1,1}]$. If $P$ is a binary matrix similar to $A$, this metric measures the number of edges in the ground truth graph that are missing or falsely added in the estimated graph; if $P$ is a matrix of link probabilities, this metric measures to what extent the links in $A$ are perturbed. It can be seen that this metric, just like GNN performance, is sensitive to link perturbations from a local perspective, and it is able to reflect the overall quality of the estimated graph topology. This metric is also closely related to the mean absolute error (MAE) between $P$ and $A$, defined as $\frac{1}{n^2}\sum_{i,j}|P_{ij}-A_{ij}|$, which is a commonly used metric in empirical evaluation. Therefore, we use the expectation of $\ell_1$-distance between $P$ and $A$ as the utility metric to quantify the utility of \textsc{Blink}, and we present an upper bound of it in the following Theorem \ref{thm:p_a_error}.

\begin{theorem}\label{thm:p_a_error}
  Assume that $\beta$ found by MLE in Algorithm \ref{alg:be} is the optimal solution that maximizes $\ell_{\tilde d^+}(\beta)$. Then we have \small$$
  \E\left[\Vert P-A\Vert_{1,1}\right]\leq2\Vert A\Vert_{1,1}+\frac{n}{2\epsilon_d},
  $$\normalsize where the expectation is taken from the randomness of RR (i.e. $\tilde A$) and Laplace mechanism (i.e. $\tilde d$).\footnote{$\Vert A\Vert_{1,1}\defeq \sum_{i=1}^n\sum_{i=1}^n |A_{ij}|$ for matrix $A\in\mathbb{R}^{n\times n}$.}
\end{theorem}

\begin{remark} [Implications of Theorem \ref{thm:p_a_error}] \label{rmk:impl1} %
  
Theorem \ref{thm:p_a_error} is significant since it shows that $P$ is a reasonable estimate of $A$ in the sense that its $\ell_1$-distance from $A$ is of the same order of magnitude as $A$ itself and $A$ is usually sparse. For example, for a random guess $P$ whose entries are all $1/2$, $\Vert P-A\Vert_{1,1}$ assumes the value $n^2/2\gg2\Vert A\Vert_{1,1}+n/(2\epsilon_d)$ when $A$ is sparse. This is reflected in Corollary \ref{col:mae} below.  Since our approach, \textsc{Blink}, is developed based on randomized response, here we compare the given bound with the estimation error of randomized response. Since the flip probability of randomized response is given as $1/(1+\exp{(\epsilon)})$, the expected estimation error for RR, in terms of the $\ell_1$ distance from $A$, is $n^2/(1+\exp{(\epsilon)})$, which is much larger than the bound given in Theorem \ref{thm:p_a_error} for sparse graphs. This shows that our approach successfully utilizes Bayesian estimation to denoise the noisy adjacency lists perturbed by randomized response and achieves a significant improvement over na\"ive approaches.
\end{remark}

\begin{corollary}\label{col:mae} For a sparse graph where $\Vert A\Vert_{1,1}=O(n)$, the expected mean absolute error (MAE) of $P$ against $A$ is bounded by $O(1/n+1/(n\epsilon))$.
\end{corollary}

\paragraph{Empirical bound tightness} To empirically evaluate the estimation accuracy of the posterior link probabilities $P$ against the ground truth graph topology $A$, and to inspect the tightness of our upper bound on the estimation error given in Theorem \ref{thm:p_a_error}, we report the average mean absolute error (MAE) and its theoretical upper bound (as given in Theorem \ref{thm:p_a_error}) between $P$ and $A$ on four well-known graph datasets in Figure \ref{fig:mae}. Figure \ref{fig:mae} shows that in all datasets, the MAE between $P$ and the ground truth $A$ is very small, and decrease to almost zero (on the order of $10^{-6}$ when $\epsilon=8$) as the privacy budget $\epsilon$ increases. This demonstrates that the inferred link probability matrix $P$ is a close estimation of the unseen private adjacency matrix $A$, and thus can be used for GNN training. Furthermore, Figure \ref{fig:mae} reports that the upper bound of the expected MAE given by Theorem \ref{thm:p_a_error} is very close to the empirical average MAE when the total privacy budget $\epsilon$ is small. However, empirical results also suggest that the bound given in Theorem \ref{thm:p_a_error} is not tight when $\epsilon$ is large, as our upper bound converges to $2\Vert A\Vert_{1,1}$ instead of $0$ when $\epsilon\to\infty$, while the empirical estimation error converges to zero when $\epsilon$ grows larger. This has inspired us to prove Theorem \ref{thm:inf} below, which states that the estimated graph from noisy messages will be identical to the actual one when $\epsilon\to\infty$.

\begin{theorem}\label{thm:inf}
  As $P$ is a random function of the total link privacy parameter $\epsilon$, we write $P=P_\epsilon$. Then, we have $\lim_{\epsilon\to\infty} P_\epsilon=A$, i.e., when $\epsilon\to\infty$, the estimated graph from noisy messages converges to the ground truth.
\end{theorem}

\begin{remark}[Implications of Theorem \ref{thm:inf}]\label{rmk:impl2}
  Theorem \ref{thm:inf} demonstrates that when $\epsilon$ goes to infinity, the estimated graph from noisy messages will converge to the ground truth graph and hence the trained GNN will also have the same performance as its theoretical upper bound -- the performance of a GNN trained with the accurate graph topology. This is a desirable property of any differentially private mechanisms, yet not enjoyed by all existing ones. For example, LDPGen \cite{qin2017generating} clusters structurally similar nodes together and generates a synthetic graph based on noisy degree vectors via the Chung--Lu model \cite{aiello2000random}. Even when no noise is injected, the generated graph is not guaranteed to be identical to the ground truth graph since only accurate degree vectors are used to construct the graph. Theorem \ref{thm:inf} shows that \textsc{Blink} is able to achieve this desirable property, and together with Theorem \ref{thm:p_a_error}, which has been shown to be quite tight when $\epsilon$ is small, we show that the estimation error of \textsc{Blink} is well controlled for all $\epsilon$, as demonstrated empirically in Figure \ref{fig:mae}.
\end{remark}

\begin{remark}
  Note that Theorem \ref{thm:p_a_error}, Corollary \ref{col:mae} and Theorem \ref{thm:inf} are not violations to privacy. They indicate how good the server can estimate the ground truth, conditioned on the fact that the input information is theoretically guaranteed to satisfy $\epsilon$-link LDP (as shown in Theorem \ref{thm:ldp}). These are known as privacy-utility bounds, and it is standard practice in local differential privacy literature for the server to denoise the received noisy information to aggregate useful information.
\end{remark}

\begin{remark}[Limitations]\label{rmk:bound_limitations}
  Note that Theorems \ref{thm:p_a_error} and \ref{thm:inf} only capture the estimation errors of $P$, and are not direct indicators of the performance of the GNNs trained on $P$. As discussed in the beginning of this section, there is no metric that can directly reflect the performance of the trained GNNs. However, in general, the closer the estimated adjacency matrix $P$ is to the ground truth $A$, the better the GNNs trained on $P$ are expected to perform. Still, Theorem \ref{thm:p_a_error} and \ref{thm:inf} alone are not sufficient to demonstrate the superior performance of the proposed approach, \textsc{Blink}, over existing approaches. For example, for \textsc{L-DpGCN} (to be introduced in Section \ref{sec:exp-setting}), the mechanism only retains around the same number of links in the estimated graph as the ground truth graph, and hence the estimation error is approximately bounded by $2\Vert A\Vert_{1,1}$, similar to what we have proved in Theorem \ref{thm:p_a_error}. This is also the case for degree-preserving randomized response proposed in \cite{hidano2022degree}. Therefore, we provide extensive empirical evaluations of the performance of \textsc{Blink} in Section \ref{sec:experiments} and show that \textsc{Blink} outperforms existing approaches in terms of utility at the same level of privacy.
\end{remark}

\subsection{Technical novelty}\label{sec:novelty}
The split of privacy budget to be separately used for degree information and adjacency lists has appeared in the literature \cite{hidano2022degree,yelfgdpr2022}. However, in both works, the noisy degrees and adjacency lists are denoised or calibrated such that a target aggregate statistic can be estimated more accurately. \citet{hidano2022degree} uses the noisy degree to sample from the noisy adjacency lists such that the overall graph density can be preserved. \cite{yelfgdpr2022} combines two estimators of the target aggregate statistic, one from noisy degrees and the other from noisy adjacency lists, and calibrates for a better estimation for the target aggregate statistic, such as clustering coefficient and modularity. As discussed previously, guarantees on the estimation error of these aggregate statistics are not sufficient to train useful GNNs due to their sensitivity to link perturbations. In contrast, our approach, \textsc{Blink}, utilizes the noisy degree information to estimate the posterior link probabilities conditioned on the evidence of noisy RR outputs \textit{for all possible links}, via Bayesian estimation. To the best of our knowledge, this is a novel approach that has not been explored in the literature. %

\section{Experiments}\label{sec:experiments}
\subsection{Experimental settings}\label{sec:exp-setting}

\paragraph{Environment} To demonstrate the privacy-utility trade-off of our proposed mechanism, we ran extensive experiments on real-world graph datasets with state-of-the-art GNN models.\footnote{Our code can be found at \url{https://github.com/zhxchd/blink\_gnn}.} The experiments are conducted on a machine running Ubuntu 20.04 LTS, equipped with two Intel\textsuperscript{\textregistered} Xeon\textsuperscript{\textregistered} Gold 6326 CPUs, 256GB of RAM and an NVIDIA\textsuperscript{\textregistered} A100 80GB GPU. We implement our mechanism and other baseline mechanisms using the PyTorch\footnote{Available at \url{https://pytorch.org/}} and PyTorch Geometric\footnote{Available at \url{https://www.pyg.org/}} frameworks. To speed up execution, we use NVIDIA's TF32 tensor cores \cite{a100} in hyperparameter search at the slight cost of precision. All experiments other than hyperparameter grid search are done using the more precise FP32 format to maintain precision.

\paragraph{Datasets} We evaluate \textproc{Blink} and other mechanisms on real-world graph datasets. The description of the datasets is as followed:\begin{itemize}[leftmargin=*]
  \item \textit{Cora} and \textit{CiteSeer} \cite{yang2016revisiting} are two well-known citation networks commonly used for benchmarking, where each node represents a document and links represent citation relationships. Each node has a feature vector of bag-of-words and a label for category.
  \item \textit{LastFM} \cite{rozemberczki2020char} is a social network collected from music streaming service LastFM, where each node represents a user and links between them represent friendships. Each node also has a feature vector indicating the artists liked by the corresponding user and a label indicating the home country of the user.
  \item \textit{Facebook} \cite{rozemberczki2021multi} is a social network collected from Facebook, where each node represents a verified Facebook page and links indicate mutual likes. Each node is associated with a feature vector extracted from site description and a label indicating the cite category. This graph is significantly larger and more dense than the previous datasets, and hence represents the scalability and performance on larger graphs of our proposed method.
\end{itemize} Table \ref{tab:dataset} summarizes the statistics of datasets used in experiments.

\paragraph{Baselines} To better present the performance of \textsc{Blink}, we implement the following baseline mechanisms for comparison.

\begin{table}[t]
  \caption{Statistics of the graph datasets used in experiments}
  \label{tab:dataset}
  \vspace{-3mm}
  \centering
  \begin{small}
  \begin{sc}
  \begin{tabular}{lcccc}
  \toprule
  Dataset    & \#Nodes & \#Features & \#Classes & \#Edges \\
  \midrule
  Cora\cite{yang2016revisiting}      &2708&1433&7&5278\\
  CiteSeer\cite{yang2016revisiting}   &3327&3703&6&4552\\
  LastFM\cite{rozemberczki2020char} &7624&128&18&27806\\
  Facebook\cite{rozemberczki2021multi} &22470&128&4&171002\\
  \bottomrule
  \end{tabular}
    \end{sc}
  \end{small}
\end{table}

\begin{figure*}[hbt!]
  \centering
  \includegraphics{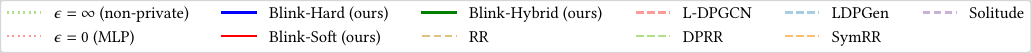}
  \vspace{5pt}

  \begin{subfigure}[t]{0.245\textwidth}
      \makebox[0pt][r]{\makebox[20pt]{\raisebox{40pt}{\rotatebox[origin=c]{90}{GCN}}}}%
      \includegraphics{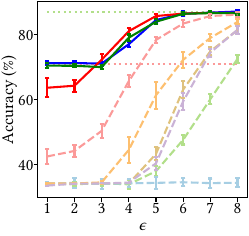}
      \makebox[0pt][r]{\makebox[20pt]{\raisebox{40pt}{\rotatebox[origin=c]{90}{GraphSage}}}}%
      \includegraphics{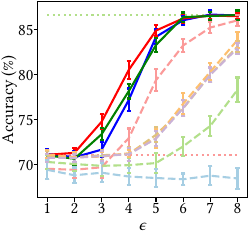}
      \makebox[0pt][r]{\makebox[20pt]{\raisebox{40pt}{\rotatebox[origin=c]{90}{GAT}}}}%
      \includegraphics{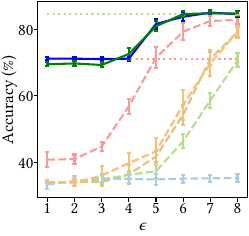}
      \caption{Cora}
  \end{subfigure}
  \begin{subfigure}[t]{0.245\textwidth}
      \includegraphics{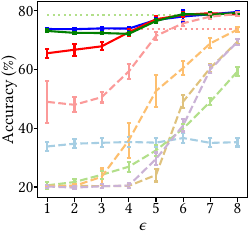}
      \includegraphics{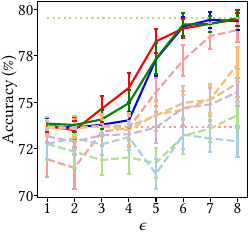}
      \includegraphics{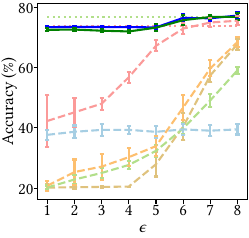}
      \caption{CiteSeer}
  \end{subfigure}
  \begin{subfigure}[t]{0.245\textwidth}
      \includegraphics{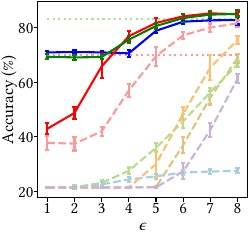}
      \includegraphics{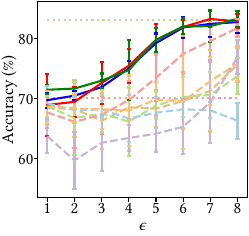}
      \includegraphics{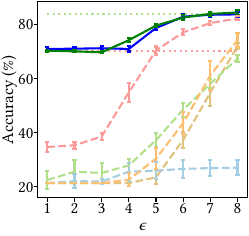}
      \caption{LastFM}
  \end{subfigure}
  \begin{subfigure}[t]{0.245\textwidth}
      \includegraphics{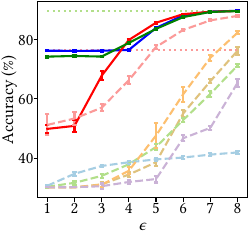}
      \includegraphics{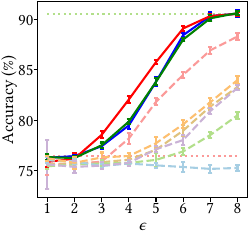}
      \includegraphics{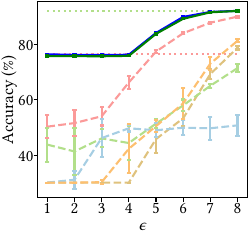}
      \caption{Facebook}
  \end{subfigure}
  \vspace{-5pt}
  \caption{Performance of \textsc{Blink} and other mechanisms. X-axis represents $\epsilon$ and y-axis represents test accuracy (\%).}
  \label{fig:fig1}
\end{figure*}

\begin{enumerate}[leftmargin=*]
  \item Randomized response (\textsc{RR}) \cite{warner1965rr} is included to demonstrate the effectiveness of our server-side denoising algorithm, where the server directly uses the RR result of adjacency matrix as the estimated graph without calibration.
  \item %
  \citet{wu2022linkteller} propose \textproc{DpGCN} as a mechanism to achieve $\epsilon$-central DP to protect graph links. It adds Laplacian noise to all entries of $A$ and keeps the top $\Vert A\Vert_{1,1}$ entries to be estimated links. However, in LDP setting, $\Vert A\Vert_{1,1}$ is kept private to the server and cannot be directly utilized. Following the same idea, we propose a LDP variant of it, namely \textproc{L-DpGCN}, where each node adds Laplacian noise to entries of its own adjacency list and sends the noisy adjacency matrix $\tilde A$ to the server. The server first estimates the number of links by $\Vert \tilde A\Vert_{1,1}$ and keeps the top $\Vert \tilde A\Vert_{1,1}$ entries as estimated links. 
  \item \textsc{Solitude} is proposed in \cite{lin2022towards} as a LDP mechanism to protect features, labels and links of the training graphs. The link LDP setting of theirs is identical to ours, and we only use the link privacy component of their mechanism. In \textsc{Solitude}, each node perturbs its adjacency list via randomized response, and the server collects noisy matrix $\tilde A$. However, RR result is usually too dense to be useful for GNN training. Hence, \textsc{Solitude} learns a more sparse adjacency matrix by replacing the original GNN learning objective with \begin{equation}\label{eq:solitude}
    \min_{\hat A, \theta}\mathcal{L}(\hat A\vert \theta)+\lambda_1\Vert \hat A - \tilde A\Vert_F+\lambda_2\Vert\hat A\Vert_{1,1},
    \end{equation} where $\theta$ is the GNN trainable parameters and $\mathcal{L}(\hat A|\theta)$ is the original GNN training loss under parameters $\theta$ and graph topology $\hat A$. To optimize for Equation (\ref{eq:solitude}), \textsc{Solitude} uses alternating optimization to optimize for both variables.
    
    \item \citet{hidano2022degree} propose DPRR (degree-preserving randomized response) to achieve $\epsilon$-link local differential privacy to train GNNs for graph classification tasks. The algorithm denoises the randomized response noisy output by sampling from links reported by RR such that the density of the sampled graph is an unbiased estimation to the ground truth density. We implement DPRR as a baseline to compare with \textsc{Blink} variants.
    \item We also implement and include baselines designed for privacy-preserving graph synthesis and analysis. \citet{qin2017generating} propose LDPGen, a mechanism to generate synthetic graphs by collecting link information from decentralized nodes with link LDP guarantees, similar to ours. The key idea of the mechanism is to cluster structually similar nodes together (via K-means \cite{arthur2007k}) and use noisy degree vectors reported by nodes to generate a synthetic graph via the Chung--Lu model of random graphs \cite{aiello2000random}.
    \item \citet{imola2021locally} propose locally differentially private mechanisms for graph analysis tasks, namely, triangle counting and $k$-star counting. Their main idea is to use the randomized response mechanism to collect noisy adjacency lists from nodes, and then derive an estimator to estimate the target graph statistics from the noisy adjacency lists. We adopt the first part of their mechanisms, i.e., random response, to derive noisy graph topology to be used for GNNs. The RR mechanism used in \cite{imola2021locally} only involves injecting noise to the lower triangular part of the adjacency matrix, i.e., node $v_i$ only perturbs $a_{i,1},\ldots,a_{i,i-1}$ and sends these bits, to force the noisy adjacency matrix to be symmetric. Hence, we denote this baseline as SymRR.
\end{enumerate} %

More discussions of these baseline methods and other related works can be found in Section \ref{sec:related}.

\paragraph{Experimental setup} For all models and datasets, we randomly split the nodes into train/validation/test nodes with the ratio of 2:1:1. To better demonstrate the performance of \textsc{Blink} and other baseline methods, we apply them with multiple state-of-the-art GNN architectures including graph convolutional networks (GCN) \cite{kipf2016semi}, GraphSAGE \cite{hamilton2017graphsage} and graph attention networks (GAT) \cite{velivckovic2017graph} (details of the model configurations can be found in Appendix \ref{sec:appendix_gnn}). Note that we do not conduct experiment on \textsc{Blink-Soft} or \textsc{Solitude} with the GAT architecture because it is not reasonable to let all nodes attend over all others \cite{velivckovic2017graph} (even in a weighted manner). To compare the performance of DP mechanisms, we also experiment on all datasets with non-private GNNs, whose performance will serve as a theoretical upper bound of all DP mechanisms. Moreover, following \cite{kolluri2022lpgnet}, we also include the performance of multi-layer perceptrons (MLPs) for each dataset, which is trained after removing all links from the graph and is considered fully link private. We experiment all mechanisms under all architectures and datasets with $\epsilon\in\{1,2,\ldots,8\}$. To showcase the full potential of our proposed method, for each combination of dataset, GNN architecture, privacy budget and mechanism, we run grid search and select the hyperparameters with the best average performance on validation data over 5 trials, and report the mean and standard deviation of model accuracy (or equivalently, micro F1-score, since each node belongs to exactly one class) on test data over 30 trials for statistical significance. Similar to previous works \cite{lin2022towards,sajadmanesh2021locally}, we do not consider the potential privacy loss during hyperparameter search. The hyperparameter spaces for all mechanisms used in grid search are described in detail in Appendix \ref{sec:app-hp}.
  
\begin{figure*}[hbt!]
  \centering
  \includegraphics{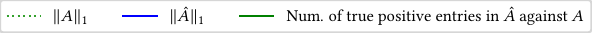}
  \vspace{5pt}

  \begin{subfigure}[b]{0.25\textwidth}
      \centering
      \includegraphics{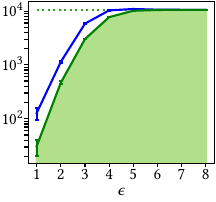}
      \vskip -0.2cm
      \caption{Cora}
      \label{fig:density_cora}
  \end{subfigure}%
  \begin{subfigure}[b]{0.25\textwidth}
      \centering
      \includegraphics{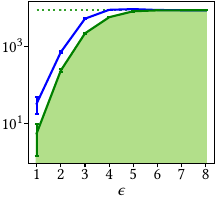}
      \vskip -0.2cm
      \caption{CiteSeer}
      \label{fig:density_citeseer}
  \end{subfigure}%
  \begin{subfigure}[b]{0.25\textwidth}
      \centering
      \includegraphics{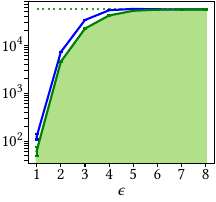}
      \vskip -0.2cm
      \caption{LastFM}
      \label{fig:density_lastfm}
  \end{subfigure}%
  \begin{subfigure}[b]{0.25\textwidth}
    \centering
    \includegraphics{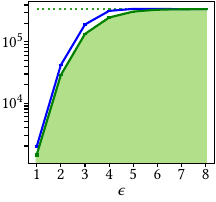}
    \vskip -0.2cm
    \caption{Facebook}
    \label{fig:density_facebook}
\end{subfigure}
  \vskip -0.3cm
  \caption{Density of estimated $\hat A$ against ground truth $A$ in \textsc{Blink-Hard}.}
  \label{fig:density}
\end{figure*}

\subsection{Experimental results and discussions}

\subsubsection{Privacy-utility of the proposed \textsc{Blink} mechanisms}

\begin{figure}[!t]
  \centering
  \includegraphics[height=26pt]{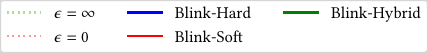}
  \vspace{5pt}
  
  \includegraphics{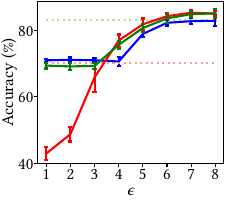}
  \vskip -0.3cm
  \caption{GCN test accuracy on LastFM with all three variants at $\epsilon\in\{1,8\}$. This is a closer look on the results in Figure \ref{fig:fig1} on LastFM with GCN.}
  \label{fig:lastfm_gcn}
\end{figure}

We report the average test accuracy of all three variants of \textsc{Blink} and other baseline methods over all datasets and GNN architectures in Figure \ref{fig:fig1}. For all methods, the test accuracy increases as the total privacy budget increases, indicating the privacy-utility trade-off commonly found in differential privacy mechanisms. At all privacy budgets, \textsc{L-DpGCN} outperforms RR because the former one takes the graph density into consideration and preserves the number of links in the estimated graph, while the latter mechanism produces too many edges after randomly flipping the bits in the adjacency matrix which renders the graph too dense. This is consistent with Remark \ref{rmk:impl1} where we show the huge improvement in estimation error for \textsc{Blink} over RR. We notice that the performance of SymRR and our implementation of \textsc{Solitude} is on par with RR, which makes sense because SymRR is essentially RR only applied to the lower triangular part of the adjacency matrix while \textsc{Solitude} denoises graph topology based on RR outputs. Note that \citet{lin2022towards} did not made the implementation of \textsc{Solitude} publicly available by the time this paper was written, and the authors only performed experiments at large privacy budgets where $\epsilon\geq 7$, while our results under similar privacy budgets agree with or outperform theirs presented in the paper. Additionally, the performance of LDPGen is worse than other mechanisms, which is expected because it is designed for graph synthesis and not for GNN training. The performance of LDPGen on GNNs also does not improve as the privacy budget increases, which is also expected because at all privacy budgets, the synthetic graph generated by LDPGen is always generated by a random graph model given noisy degree vectors. This has been discussed in Remark \ref{rmk:impl2}. 

It is evident from Figure~\ref{fig:fig1} that at all levels of $\epsilon$, the \textsc{Blink} variants generally outperform all baseline methods, because they also take individual degrees into consideration when estimating the prior probabilities, and utilize its confidence in links via hard thresholding, soft weighted aggregation or both. Noticeably, only \textsc{Blink} variants (especially \textsc{Blink-Hard} and \textsc{Blink-Hybrid}) can consistently perform on par with the fully link private MLP baselines, which is due to the fact that these variants can eliminate noisy and non-confident link predictions at low privacy budget and high uncertainty. Additionally, among the three GNN architectures, the baseline methods can perform better on GraphSAGE with accuracy closer to MLP. This is because GraphSAGE convolutional layers have a separate weight matrix to transform the embedding of the root node and hence can learn not to be distracted by the embeddings of false positive neighbors. See Appendix \ref{sec:appendix_gnn} for details on GNN architectures. At last, for $\epsilon\in[4,8]$ which is widely adopted in LDP settings in real-world industry practice \cite{appledpoverview,apple2017learning}, Blink variants on different GNN architectures outperform the MLP and baselines significantly in most cases, indicating their utility under real-world scenarios. Also, when $\epsilon \geq 6$, Blink variants achieve test accuracy on par with the theoretical upper bound on all datasets and architectures. In the following paragraphs, we describe with greater detail the performance and trade-off among the \textsc{Blink} variants.

\paragraph{Performance of \textsc{Blink-Hard}} As demonstrated in Figure \ref{fig:fig1}, one main advantage of \textsc{Blink-Hard} is that it is almost never outperformed by MLP trained only on node features, which is not the case for the baseline methods. Existing approaches \cite{wu2022linkteller,hidano2022degree} of achieving (central or local) link privacy on graphs aim to preserve the graph density in the estimated graph, i.e. try to make $\Vert \hat A\Vert_{1,1}\approx\Vert A\Vert_{1,1}$, however, when $\epsilon$ is small, it is against the promise of differential privacy to identify the same number of links from the estimated graph from the actual graph. As \citet{kolluri2022lpgnet} point out, $100\%$ of the selected top $|E|$ links estimated by \textsc{DpGCN} \cite{wu2022linkteller} at $\epsilon=1$ are false positive, corrupting the GNN results when aggregating neighbor embeddings. By only keeping links whose posterior link probability of existence exceeds $0.5$, \textsc{Blink-Hard} takes an alternative approach of understanding graph density at tight privacy budgets and high uncertainty. As shown in Figure \ref{fig:density}, $\Vert \hat A \Vert_{1,1}$ estimated by \textsc{Blink-Hard} is much lower than the ground truth density $\Vert A\Vert_{1,1}$ when $\epsilon$ is small, and gradually increases to a similar level with $\Vert A\Vert_{1,1}$ as $\epsilon$ increases. In this way, \textsc{Blink-Hard} eliminates information that is too noisy to be useful at low privacy budgets, thus reducing false positive link estimations and avoiding them from corrupting the GNN model. As shown in Figure \ref{fig:density}, among the much fewer link estimations given by $\textsc{Blink-Hard}$, the true positive rates are much higher than \textsc{DpGCN} as reported in \cite{kolluri2022lpgnet}. Therefore, \textproc{Blink-Hard} consistently outperforms the fully link-private MLP and other baselines.

\paragraph{Performance of \textsc{Blink-Soft}} Although \textsc{Blink-Hard} outperforms MLP and baseline mechanisms, the elimination and rounding of link probabilities lead to a significant amount of information loss. \textsc{Blink-Soft} aims to improve over the hard variant at moderate privacy budgets by utilizing the extra information while it is not too noisy. As described in Section \ref{sec:variants}, \textsc{Blink-Soft} utilizes the inferred link probabilities by using them as weights in the GNN aggregation step (see Appendix \ref{sec:appendix_gnn} for more details), which has enabled the GNN to be fed with more information and perform better as long as the extra information is useful. As reflected in Figure \ref{fig:fig1} and Figure \ref{fig:lastfm_gcn}, \textsc{Blink-Soft} is able to outperform \textsc{Blink-Hard} at moderate privacy budgets (i.e., $\epsilon\in[4,6]$) under almost all dataset and GNN architecture combinations. For higher privacy budgets, as both variants perform very well and are on par with the non-private upper bound, the performance gap is not significant. However, at lower privacy budgets where $\epsilon\in[1,3]$, \textsc{Blink-Soft} sometimes performs much worse than \textsc{Blink-Hard} and the fully private MLP baseline, for example, on LastFM with GCN model which we take a closer look in Figure \ref{fig:lastfm_gcn}. This is caused by the low information-to-noise ratio of the inferred link probabilities at low privacy budgets. Here, we confirm the hypothesis proposed in Section \ref{sec:variants} that \textsc{Blink-Hard} and \textsc{Blink-Soft} complement each other where the hard variant performs better at low privacy budget while the soft variant performs better at higher privacy budgets.

\paragraph{Performance of \textsc{Blink-Hybrid}} The hybrid variant is proposed to combine the previous two aiming to enjoy the benefits of both variants across all privacy settings. As shown in Figure \ref{fig:fig1}, at low privacy budgets, \textsc{Blink-Hybrid} successfully outperforms \textsc{Blink-Soft} by a significant margin and achieves test accuracy on par with \textsc{Blink-Hard} (thus, not outperformed by MLP), due to its elimination of noisy and useless information, avoiding false positive links from poisoning the model. At higher privacy budgets, \textsc{Blink-Hybrid} is often able to perform better than the hard variant, thanks to keeping the link probabilities as aggregation weights. For example, for the configuration of LastFM with GCN in Figure \ref{fig:fig1} which we take a closer look in Figure \ref{fig:lastfm_gcn}, \textsc{Blink-Hard} achieves accuracy close to \textsc{Blink-Hard} at $\epsilon\in[1,3]$ while performs on par with \textsc{Blink-Soft} at $\epsilon\in[4,8]$, achieving the best of both worlds. Although \textsc{Blink-Hybrid} is seldom able to outperform both the hard and the soft variants, it can enjoy both the benefits of \textsc{Blink-Hard} at low privacy budgets and the benefits of \textsc{Blink-Soft} at higher privacy budgets.

\begin{figure}[!t]
  \centering
  \includegraphics[height=13pt]{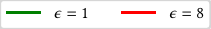}
  \vspace{5pt}
  
  \includegraphics{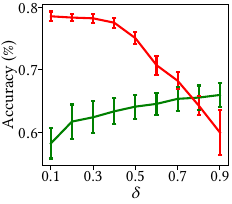}
  \vskip -0.3cm
  \caption{GCN test accuracy on CiteSeer with \textsc{Blink-Soft} at $\epsilon\in\{1,8\}$.}
  \label{fig:delta}
\end{figure}

\subsubsection{On the effects of $\delta$}\label{sec:delta}

The degree privacy budget parameter, $\delta$, is an important hyperparameter that makes a difference on the performance of the trained GNNs. In previous experiments, we choose the value of $\delta$ by grid search and use the one that is associated with the best validation accuracy. To better understand the effects of different choices of $\delta$ values on the GNN performance, we report the test accuracy of graph convolutional network on CiteSeer with \textsc{Blink-Soft} over varying $\delta$ values at $\epsilon\in\{1,8\}$ in Figure \ref{fig:delta}. It can be seen that at different privacy budgets, there are different implications for the values of $\delta$. At small privacy budget, i.e., $\epsilon=1$, we notice that the GNN performance increases as $\delta$ value increases, while at larger privacy budget, e.g., $\epsilon=8$, it is clear that lower $\delta$ values result in better performance. This is because at very tight privacy budgets, the noisy adjacency matrix given by randomized response will be too noisy to be useful, hence, the prior probabilities estimated from the noisy degree sequence will be more important. Therefore, it is optimal to allocate more privacy budget towards degrees. At much higher privacy budgets, the flip probability in randomized response becomes so small that the noisy adjacency matrix itself is sufficient to provide the necessary information to effectively train the GNN, hence, it is preferred to allocate more privacy budget to the adjacency lists. %
If we denote $\delta^*(\epsilon)$ as the optimal $\delta$ at total privacy budget $\epsilon$ that achieves the best performing GNNs (for instance, we have $\delta^*(1)=0.9$ and $\delta^*(8)=0.1$ in Figure \ref{fig:delta}), we propose a conjecture that $\delta^*$ decreases as $\epsilon$ increases, i.e., ${\delta^{*}}'(\epsilon)<0$, and our experiments resonate with this conjecture.

\begin{figure}[!t]
  \centering
  \includegraphics[height=13pt]{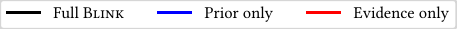}
  \vspace{5pt}

  \includegraphics{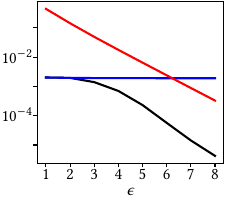}%
  \includegraphics{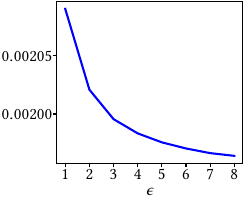}
  \vskip -0.3cm
  \caption{The MAE of the estimated link probabilities $P$ against ground truth $A$ for full \textsc{Blink}, \textsc{Blink} with prior component only and \textsc{Blink} with evidence component only on CiteSeer. The latter figure is a closer look at the prior component whose trend is unclear in the former figure.}
  \label{fig:ablation}
\end{figure}

\subsubsection{Ablation studies}

Naturally, one would be curious about which component, the prior or the evidence, of the proposed \textsc{Blink} mechanisms, contributes more to the final link estimations $P$. To answer this question, we conduct ablation studies on the proposed methods. Figure \ref{fig:ablation} reports the MAE of the estimated link probabilities $P$ against the ground truth $A$ under \textsc{Blink}, its prior component and its evidence components. \textsc{Blink} with prior component only is equivalent to taking $\delta=1$ where the flip probability of RR becomes $1/2$ and hence the noisy adjacency matrix does not provide any information as evidence. \textsc{Blink} with evidence only is the case where the prior probabilities are set to be all $1/2$ to provide no extra information. First, as shown in Figure \ref{fig:ablation}, all mechanisms, the complete and partial ones, have their MAE decreasing as privacy budgets increase. More importantly, it can be seen that at tighter privacy budgets, the prior-only mechanism produces better estimations than its evidence-only counterpart, indicating that the prior contributes more to the final estimation at tighter privacy budgets. When $\epsilon$ grows (i.e., $\epsilon\geq7$ in Figure \ref{fig:ablation}), the noisy adjacency matrix becomes less noisy so the evidence-only method starts to produce better estimations, playing a more important role than the prior. This agrees with the findings in Section \ref{sec:delta} that it is optimal to allocate more privacy budget to degrees (i.e., the prior component) at smaller $\epsilon$ and vise versa. It is important to note that at all privacy budgets, the full \textsc{Blink} method significantly outperforms both single-component methods, indicating that our proposed method effectively utilizes both components to make better estimations and both components are irreplaceable in \textsc{Blink}.

\section{Related work} \label{sec:related}

\paragraph{Graph neural networks} Recent years have witnessed an emerging amount of work on graph neural networks for many tasks on graph, such as node classification, link classification and graph classification. Many novel GNN models have been proposed, including GCN \cite{kipf2016semi}, GraphSAGE \cite{hamilton2017graphsage}, GAT \cite{velivckovic2017graph} and Graph Isomorphism Networks \cite{xu2018powerful}. As our proposed mechanism estimates the graph topology and then feed it into GNN models without interfering the model architecture, we will not survey recent advances in GNNs here in great detail, but refer the audience to available surveys \cite{wu2020comprehensive, bronstein2017geometric, ZHOU202057} for detailed discussions of GNN models, performance and applications. 

\paragraph{Differentially private GNNs} There have been recent attempts in the literature of incorporating the notion of differential privacy to GNNs. \citet{wu2022linkteller} study the adversarial link inference attacks on GNNs and proposes \textsc{DpGCN}, a central DP mechanism to protect edge-level privacy, which can be easily modified to adopt stronger LDP guarantee. \citet{daigavane2021node} attempt to extend the well-celebrated DP-SGD \cite{abadi2016deep} algorithm on neural networks to GNNs and achieve stronger node-level central differential privacy. More recently, \citet{kolluri2022lpgnet} propose new GNN architecture to achieve edge-level central DP, where they separate the edge structure and only use MLPs to model both node features and graph structure information. Following a similar intuition, \citet{sajadmanesh2022gap} propose a new mechanism where the aggregation step is decoupled from the GNN and executed as a pre-processing step to save privacy budget. When combined with DP-SGD, \citet{sajadmanesh2022gap} achieve stronger node-level central DP on the altered GNN architecture.

For local differential privacy, \citet{sajadmanesh2021locally} propose a LDP mechanism to protect node features but not the graph topology. \citet{lin2022towards} extend on \cite{sajadmanesh2021locally} and propose \textproc{Solitude} to also protect edge information in a LDP setting. The link LDP notion of \cite{lin2022towards} is identical to that of ours. However, their link DP mechanism is not principled and their estimated graph structure is learned by minimizing a loss function $\Vert \hat A-\tilde A\Vert_{1,1}+\lambda\Vert\hat A\Vert_F$ to encourage the model to choose less dense graphs. \citet{hidano2022degree} propose link LDP mechanism for graph classification tasks, and takes a similar approach to ours, by separately injecting noise to adjacency matrix and degrees. However, \citet{hidano2022degree} aim to preserve node degrees in the estimated graph like \textsc{DpGCN}, which is not suitable to node classification tasks and performs worse than our method as shown in Section \ref{sec:experiments}.

\paragraph{Privacy-preserving graph synthesis} Privacy-prserving \allowbreak graph publication is also closely related to what we have studied, where one aims to publish a sanitized and privacy-preserving graph given an input graph. \citet{blocki2012johnson} utilize the Johnson-Lindenstrauss Transform to achieve graph publication with edge differential privacy. \citet{qin2017generating} consider local edge differential privacy where an untrusted data curator collects information from each individual user about their adjacency lists and construct a representative synthetic graph of the underlying ground truth graph with edge LDP guarantee. This is achieved by incrementally clustering structurally similar users together. More recently, \citet{yang2021secure} achieve differentially private graph generation by noise injection to a graph generative adversarial network (GAN) such that the output of the GAN model is privacy-preserving. It is worth noting that in the settings of \cite{blocki2012johnson,yang2021secure}, a privacy-preserving synthetic graph is generated in a centralized way, i.e., the curator has access to the ground truth graph and perturbs it for a privacy-preserving publication, which is a weaker threat model than ours. \citet{qin2017generating} consider threat models similar to ours with local differential privacy where the curator does not need to have access to the actual graph, but there's no theoretical upper bound on the distance from the synthetic graph to the ground truth graph, which we provide in Theorem \ref{thm:p_a_error}.

\paragraph{Privacy-preserving graph analysis} There exist prior works in the literature proposed for graph analysis tasks with local differential privacy. \citet{imola2021locally} propose mechanisms to derive estimators for triangle count and $k$-star count in graphs with link LDP. \citet{yelfgdpr2022} propose a general framework for graph analysis with local differential privacy, to estimate an arbitrary aggregate graph statistic, such as clustering coefficient or modularity. This approach combines two estimators of the target aggregate statistic, one from noisy neighbor lists and one from noisy degrees, and derives a better estimator for the target statistic. However, this approach does not produce an estimated graph topology that can be used for GNN training. Hence, we do not include this as a baseline in our experiments in Section \ref{sec:experiments}.

\paragraph{Link inference attacks in GNNs} As the popularization of GNNs in research and practice in recent years, there has garnered an increasing amount of attention on their privacy and security in the research community, and several privacy attacks on GNNs have been proposed for an attacker to infer links in the training graph. \citet{he2021stealing} propose multiple link stealing attacks for an adversary to infer links in the training graph given black-box access to the trained GNN model, guided by the heuristic that two nodes are more likely to be linked if they share more similar attributes or embeddings. \citet{wu2022linkteller} consider a scenario where a server with full access to graph topology trains a GNN by querying node features and labels from node clients (who do not host graph topology), and demonstrate that the nodes can infer the links from the server by designing adversarial queries via influence analysis. \citet{zhang2022inference} propose graph reconstruction attacks where an adversary examines the trained graph embeddings and aims to reconstruct a graph similar to the ground truth graph used in GNN training. All these attacks share the same threat model where the GNN is trained with complete and accurate information and an adversary aims to infer links by examining the trained model. Our proposed solution, \textsc{Blink}, naturally defends this kind of attacks at its source as a local differential privacy mechanism with a more severe threat model, where even the server who trains the GNN does not have non-private access to any links in the training graph.

\paragraph{Estimate of link probability given degree sequence}
The model and estimation of random graphs given degree sequence is a common topic in network science and probability theory. \citet{chatterjee2011random} discuss the maximum likelihood estimate of parameters in $\beta$-model, which is closely related to BTL model for ranking \cite{blt1952,blt1959}. Parameters in BTL model can be estimated via MM algorithms \cite{hunter2004mm} or MLE \cite{simons1999asymptotics}. Alternatively, configuration model \cite{newman2018onciguration} can also be used to model random graphs given degree sequence, which generates multi-graphs that allows multiple edges between two vertices. In configuration model, the expected number of edges between two nodes $v_i$ and $v_j$ conditioned on the degree sequence $d$ is given by $d_i d_j/(\sum d_i -1)$. When this value $\ll1$, it can be considered a probability that there's (at least one) edge between $v_i$ and $v_j$. We also attempt \textproc{Blink} with configuration model instead of $\beta$-model, but the link probabilities fail to be consistently below $1$.

\section{Conclusion}\label{sec:conclusion}
Overall, the presented framework, \textsc{Blink}, is a step towards making GNNs locally privacy-preserving while retaining their accuracy. %
It separately injects noise to adjacency lists and node degrees, and uses the latter as prior and the former as evidence to evaluate posterior link probabilities to estimate the ground truth graph. We propose three variants of \textsc{Blink} based on different approaches of constructing the graph estimation from the inferred link probabilities. Theoretical and empirical evidence support the state-of-the-art performance of \textsc{Blink} against existing link LDP approaches.

The area of differentially private GNNs is still novel with many open challenges and potential directions. There are a few future research directions and improvements for the presented paper. First, one may want to improve the bound in Theorem \ref{thm:p_a_error}, which would require careful inspection of $\hat \beta$ found by MLE. Also, an interesting future direction of research is to design algorithms such that each node can optimally decide its own privacy parameter $\delta$, to avoid the use of hyperparameter search of $\delta$ which may potentially lead to information leakage \cite{papernot2022hyperparameter}. We also leave the investigation of such potential risk to future work. Additionally, one could consider exploring different models for graph generation from the posterior link probabilities, or extend the proposed framework to other types of graphs such as directed or weighted graphs. Furthermore, exploring the scalability of \textsc{Blink} for large-scale graph data is an important future direction. At last, one may also want to incorporate other LDP mechanisms that protect features and labels (such as \cite{sajadmanesh2021locally}) into \textsc{Blink} to provide complete local privacy protection over decentralized nodes. 

\section*{Acknowledgements}

This work is supported by the Singapore Ministry of Education Academic Research Fund (AcRF) Tier 3 under MOE's official grant number MOE2017-T3-1-007 and AcRF Tier 1 under grant numbers A-8000980-00-00 and A-8000189-01-00.

\bibliographystyle{ACM-Reference-Format}
\balance
\bibliography{ref}

\newpage\appendix\label{sec:appendix}
\section{Complete proofs}\label{sec:proof}
\subsection{Proof of Theorem \ref{thm:ldp}}
\begin{proof}
  Let $\epsilon$ be the privacy budget and $\delta$ be the degree privacy parameter. Assume $a,a'\in\{0,1\}^n$ are adjacency lists that differ only at the $j$-th bit, and $(o_1,o_2)\in\{0,1\}^n\times\mathbb{R}$ is an arbitrary output of function \textproc{LinkLDP} described in Algorithm \ref{alg:nodeldp}. For ease of  presentation, we denote the function \textproc{LinkLDP} as mechanism $\mathcal{M}$, the process of randomized response in Lines 4-6 in Algorithm \ref{alg:nodeldp} as mechanism $\mathcal{L}$, and the process of adding Laplacian noise to $d$ in Lines 7-9 in Algorithm \ref{alg:nodeldp} as mechanism $\mathcal{R}$. Let $\mathcal{R}(a)=b$, $\mathcal{R}(a')=b'$, $d=\sum_{i=1}^n a_i$ and $d'=\sum_{i=1}^n a'_i$. Then
  \begin{align}
    &\frac{\Pr[\mathcal{M}(a)=(o_1,o_2)]}{\Pr[\mathcal{M}(a')=(o_1,o_2)]}=\frac{\Pr[\mathcal{R}(a)=o_1]\Pr[\mathcal{L}(d)=o_2]}{\Pr[\mathcal{R}(a')=o_1]\Pr[\mathcal{L}(d')=o_2]}\label{10}%
    \\
    &= \frac{\prod_{i=1}^n \Pr[b_i=o_{1i}]}{\prod_{i=1}^n \Pr[b'_i=o_{1i}]}\cdot\frac{\epsilon_d/2\exp(-\epsilon_d|d-o_2|)}{\epsilon_d/2\exp(-\epsilon_d|d'-o_2|)}\label{11}%
    \\
    &=\frac{\Pr[b_j=o_{1i}]}{\Pr[b'_j=o_{1i}]}\cdot \exp(-\epsilon_d(|d-o_2|-|d'-o_2|))\label{12}%
    \\
    &\leq \frac{{\exp(\epsilon_a)}/(1+\exp(\epsilon_a))}{{1}/(1+\exp(\epsilon_a))}\cdot\exp(\epsilon_d|(d-o_2)-(d'-o_2)|)\label{13}%
    \\
    &=\exp(\epsilon_a)\exp(\epsilon_d)\label{14}%
    \\
    &= \exp(\epsilon_a+\epsilon_d)=\exp(\epsilon),\label{15}
  \end{align} where (\ref{10}) and (\ref{11}) are due to independence, (\ref{12}) is because $b_i=b'_i$ when $i\neq j$ and (\ref{13}) is the result of RR and the triangle inequality.
\end{proof}

\subsection{Proof of Theorem \ref{thm:mle}}\label{proof:mle}
\begin{proof}
  Fix sequence $(\beta_i)$, we have
  \begin{gather*}
  \ell_d(\beta)=\sum_i \beta_i d_i -\log\left[\prod_{i>j}(1+\exp(\beta_i+\beta_j))\right]\\
  \ell_{\tilde d}(\beta)=\sum_i \beta_i \tilde d_i -\log\left[\prod_{i>j}(1+\exp(\beta_i+\beta_j))\right]\\
  \E\left[\ell_{\tilde d}(\beta)\right]=\ell_d(\beta)\\
  \Var\left[\ell_{\tilde d}(\beta)\right]=\Var\left[\sum_i \beta_i d_i\right]=\frac{2\sum_i\beta_i^2}{\epsilon_d^2}\end{gather*}
  Hence, due to Chebyshev's inequality, we have
  \begin{gather*}\forall a>0, \Pr\left[\left|\ell_{\tilde d}(\beta)-\ell_d(\beta)\right|>a\sqrt{\frac{2\sum_i\beta_i^2}{\epsilon_d^2}}\right]<\frac{1}{a^2}\\
  \therefore\Pr\left[\ell_{\tilde d}(\beta)-\ell_d(\beta)>b\sqrt{\frac{\sum_i\beta_i^2}{\epsilon_d^2}}\right]<\frac{1}{b^2}\quad\text{(symmetry, let }b=\sqrt{2}a\text{)}
  \end{gather*}
  With probability at least $1-{1}/{b^2}$, we have $$\displaystyle\ell_d(\beta)>\ell_{\tilde d}(\beta)-b\sqrt{\frac{\sum_i\beta_i^2}{\epsilon_d^2}}\geq \ell_{\tilde d}(\beta)-\frac{bM\sqrt{n}}{\epsilon_d^2}.$$
\end{proof}

\subsection{Proof of Theorem \ref{thm:p_a_error}}
\begin{proof}
\newcommand{\pf}{f}
\begin{align*}
  \E\left[\Vert P-A\Vert_{1,1}\right]&=\E\left[\sum_{i=1}^n\sum_{j=1}^n |P_{ij}-A_{ij}|\right]
  \\
  &=\sum_{i=1}^n\sum_{j=1}^n \E\left[|P_{ij}-A_{ij}|\right].
\end{align*}

Note that the expectation is taken over the randomness of both $\tilde A$ and $\tilde d$. For now, we only consider the expectation over the randomness of $\tilde A$ and assume that $\tilde d$ is fixed (as a result, $p_{ij}$ are fixed), until further specified.

For a particular pair $(i,j)$, we consider two cases based on $A_{ij}$. In this proof, to make the notations more concise, we denote $p_{\text{flip}}$ as $f$.

First, when $A_{ij}=0$, $|P_{ij}-A_{ij}|=P_{ij}$. As per Algorithm \ref{alg:be}, $P_{ij}=q_{ij}p_{ij}/(q_{ij}p_{ij}+q'_{ij}(1-p_{ij}))$ where $q_{ij},q'_{ij}$ are determined by random variables $\tilde A_{ij}, \tilde A_{ji}$. Therefore, we have
  $$\begin{aligned}
  &\E\left[P_{ij}\right]=\pf^2\frac{(1-\pf)^2p_{ij}}{(1-\pf)^2p_{ij}+\pf^2(1-p_{ij})}\\
  &\qquad+(1-\pf)^2\frac{\pf^2p_{ij}}{\pf^2p_{ij}+(1-\pf)^2(1-p_{ij})}+(2\pf-2\pf^2)p_{ij}\\
  &=p_{ij}\left[\left(\frac{1}{(1-2\pf)p_{ij}+\pf^2} + \frac{1}{(2f-1)p_{ij}+(1-f)^2}\right)\pf^2(1-\pf)^2\right]\\
  &\qquad+p_{ij}(2f-2f^2).
  \end{aligned}$$
  Let
  \begin{gather*}
    u_1^2=\frac{1}{(1-2\pf)p_{ij}+\pf^2},
    u_2^2=\frac{1}{(2f-1)p_{ij}+(1-f)^2},\\
    v_1^2=(1-2\pf)p_{ij}+\pf^2,
    v_2^2=(2f-1)p_{ij}+(1-f)^2.
  \end{gather*}
  As $0<f<1/2$ and $0<p_{ij}<1$, we have \begin{gather*}
    \frac{1}{(1-f)^2}<\frac{1}{(1-2\pf)p_{ij}+\pf^2}<\frac{1}{f^2},\\
    \frac{1}{(1-f)^2}<\frac{1}{(2f-1)p_{ij}+(1-f)^2}<\frac{1}{f^2},\\
    f^2<(1-2\pf)p_{ij}+\pf^2<(1-f)^2,\\
    f^2<(2f-1)p_{ij}+(1-f)^2<(1-f)^2.
  \end{gather*}
  Therefore, we have $1/(1-f)=m_1<u_1,u_2<M_1=1/f$ and $f=m_2<v_1,v_2<M_2=1-f$, according to the P{\'o}lya--Szeg{\"o}'s inequality \cite{polya1951isoperimetric}, we have
  \begin{align*}
    &(u_1^2+u_2^2)(v_1^2+v_2^2)\\
    \leq&\frac{1}{4}\left(\sqrt{\frac{M_1M_2}{m_1m_2}}+\sqrt{\frac{m_1m_2}{M_1M_2}}\right)^2(u_1v_1+u_2v_2)^2\\
    =&\frac{(f^2+(1-f)^2)^2}{f^2(1-f)^2}.
  \end{align*}
  Hence, $$
  \frac{1}{(1-2\pf)p_{ij}+\pf^2} + \frac{1}{(2f-1)p_{ij}+(1-f)^2}\leq \frac{f^2+(1-f)^2}{f^2(1-f)^2}.
  $$
  Substituting back, we have that when $A_{ij}=0$, \begin{equation}\label{16}
  \E\left[|P_{ij}-A_{ij}|\right]\leq (f^2+(1-f)^2 + 2f-2f^2)p_{ij} = p_{ij}.
\end{equation}

Second, when $A_{ij}=1$, $|P_{ij}-A_{ij}|=1-P_{ij}$. Again, we have
$$\begin{aligned}
  &\E\left[P_{ij}\right]=f^2\frac{f^2p_{ij}}{f^2p_{ij}+(1-f)^2(1-p_{ij})}\\
  &\qquad+(1-\pf)^2\frac{(1-\pf)^2p_{ij}}{(1-\pf)^2p_{ij}+\pf^2(1-p_{ij})}+(2\pf-2\pf^2)p_{ij}\\
  =&p_{ij}\left[\frac{f^4}{f^2p_{ij}+(1-f)^2(1-p_{ij})} + \frac{(1-\pf)^4}{(1-\pf)^2p_{ij}+\pf^2(1-p_{ij})}\right]\\
  &\qquad+p_{ij}(2f-2f^2).
\end{aligned}$$
According to the Cauchy--Schwarz inequality, we have $$
\begin{aligned}
&\left(\frac{f^4}{f^2p_{ij}+(1-f)^2(1-p_{ij})} + \frac{(1-\pf)^4}{(1-\pf)^2p_{ij}+\pf^2(1-p_{ij})}\right)\\
&\qquad \cdot\left(f^2p_{ij}+(1-f)^2(1-p_{ij})+(1-\pf)^2p_{ij}+\pf^2(1-p_{ij})\right)\\
=&\left(\frac{f^4}{f^2p_{ij}+(1-f)^2(1-p_{ij})} + \frac{(1-\pf)^4}{(1-\pf)^2p_{ij}+\pf^2(1-p_{ij})}\right)\\
&\qquad\cdot (f^2+(1-f)^2)\\
\geq&(f^2+(1-f)^2)^2.
\end{aligned}
$$
Thus, we have $\E[P_{ij}]\geq (f^2+(1-f)^2+2f-2f^2)p_{ij}=p_{ij}$, so when $A_{ij}=1$, 
\begin{equation}\label{17}\E[|P_{ij}-A_{ij}|]=1-\E[P_{ij}]\leq 1-p_{ij}.\end{equation}

Combining Equation (\ref{16}) and (\ref{17}), we have the following
\begin{align*}
  \E\left[\Vert P-A\Vert_{1,1}\right]&\leq\sum_{\substack{i,j \\ A_{ij}=0}}p_{ij}+\sum_{\substack{i,j \\ A_{ij}=1}}(1-p_{ij})\\
  &=\Vert A\Vert_{1,1}+\sum_{\substack{i,j \\ A_{ij}=0}}p_{ij}-\sum_{\substack{i,j \\ A_{ij}=1}}p_{ij}\\
  &\leq \Vert A\Vert_{1,1}+\sum_{i,j}p_{ij}.
\end{align*}%

Note that up to this point, we have only taken the expectation over the randomness of $\tilde A$ instead of $\tilde d$, and treat $p_{ij}$ as fixed. Now we take the randomness of $\tilde d$ into consideration.

First, due to the assumption that $\beta$ found by MLE is the optimal solution that maximizes $\ell_{\tilde d^+}(\beta)$, we have $\sum_{i,j}p_{ij}=\sum_i \tilde d_i^+\leq \sum_i (d_i+l_i^+)$, where the last inequality is due to the clipping of $\tilde d$. For each $i$, $$\E[l_i^+]=\int_0^\infty \frac{\epsilon_d}{2}\exp(-\epsilon_dx)x\, \mathrm{d}x=\frac1{2\epsilon_d},$$ since $l_i\sim\text{Laplace}(0,1/\epsilon_d)$.

Therefore, we have the final result $$
\begin{aligned}
\E[\Vert P-A\Vert_{1,1}]&\leq \Vert A\Vert_{1,1}+\sum_{i,j}\E[p_{ij}]\\
&\leq \Vert A\Vert_{1,1} + \sum_{i}\left(d_i+\E[l_i^+]\right)\\
&=2\Vert A\Vert_{1,1} + \frac{n}{2\epsilon_d}.
\end{aligned}
$$
\end{proof}

\subsection{Proof of Theorem \ref{thm:inf}}
\begin{proof}
Since $$\lim_{\epsilon\to\infty}p_\text{flip}=\lim_{\epsilon\to\infty}\frac{1}{1+\exp((1-\delta)\epsilon)}=0,$$ the noisy adjacency matrix given by RR $\tilde A$ will be identical to $A$. Therefore, when $\epsilon\to\infty$, for any $1\leq i\neq j\leq n$, \begin{itemize}[leftmargin=*]\item if $(\tilde A_{ij},\tilde A_{ji})=(A_{ij},A_{ji})=(0,0)$, $q_{ij}\to0$ and $q'_{ij}\to1$, we have $P_{ij}\to0=A_{ij}$;
    \item if $(\tilde A_{ij},\tilde A_{ji})=(A_{ij},A_{ji})=(1,1)$, $q_{ij}\to1$ and $q'_{ij}\to0$, we have $P_{ij}\to1=A_{ij}$.
  \end{itemize} Therefore, $\lim_{\epsilon\to\infty}P_\epsilon=A$.
\end{proof}

\section{Experimental details}\label{sec:app-experiments}

\subsection{GNN architectures}\label{sec:appendix_gnn}

For all GNN architectures we experiment on, we have the same model structure of a convolutional layer with 16 units, followed by a ReLU operator and a dropout layer with dropout rate set to $p_\text{dropout}$, and finally followed by another convolutional layer whose number of units equal to the number of classes of the input graph.

We experiment with different convolutional layers, including GCN, GraphSAGE and GAT, which are described in more detail below.

\paragraph{GCN} In a graph convolutional network \cite{kipf2016semi}, the node embedding is updated through \begin{equation}\label{eq:gcn}
  x_i^{(k)}=W\cdot\sum_{v_j\in\mathcal{N}(v_i)\cup \{v_i\}}\frac{1}{\sqrt{(1+\text{deg}(v_j))(1+\text{deg}(v_i))}}x_j^{(k-1)}+b,
\end{equation} where $W$ is a learnable weight matrix and $b$ is a learnable additive bias. Note that we include self-loops and symmetric normalization coefficients such that the new node embedding will depend on the previous embedding of itself.

In a weighted graph setting with edge weight matrix $P$, we use \begin{equation}\label{eq:gcn-p}
  x_i^{(k)}=W\cdot\sum_{j=1}^n\hat P_{ij}\cdot \frac{1}{\sqrt{\sum_{k=1}^n \hat P_{jk}\sum_{k=1}^n \hat P_{ik}}}x_j^{(k-1)}+b,
\end{equation} where $\hat P=P+I$ for adding self loops to the GNN.

\paragraph{GraphSAGE} In GraphSAGE \cite{hamilton2017graphsage}, the node embedding is updated through \begin{equation}
x_i^{(k)} = W_1 x_i^{(k-1)} + W_2 \cdot \frac{1}{|\mathcal{N}(v_i)|}\sum_{v_j\in\mathcal{N}(v_i)} x_j^{(k-1)} + b,
\end{equation} where $W_1,W_2$ are learnable weight matrices and $b$ is a learnable additive bias. The key difference to our configuration of GCN in Eq. (\ref{eq:gcn}) is that the transformation of the root node embedding $x_i$ is now learned separately compared to its neighbors. This enables GraphSAGE to prefer the root node embedding more than that of neighbors to achieve better performance when the links are noisy and neighbors may not have a positive effect on model performance.

In a weighted graph setting with edge weight matrix $P$, we use \begin{equation}
  x_i^{(k)} = W_1 x_i^{(k-1)} + W_2 \cdot \frac{1}{\sum_{j=1}^nP_{ij}}\sum_{j=1}^n P_{ij} x_j^{(k-1)} + b.
  \end{equation}

\paragraph{GAT} In graph attention network \cite{velivckovic2017graph}, the node embedding is updated through\begin{gather}
  x_i^{(k)} = W\cdot\sum_{v_j\in\mathcal{N}(v_i)\cup \{v_i\}}\alpha_{i,j} x_j^{(k-1)}+b,\\
 \alpha_{i,j} =\frac{\exp(\text{LeakyReLU}(a^T[Wx_i^{(k-1)}\Vert Wx_j^{(k-1)}]))}{\displaystyle\sum_{v_t\in\mathcal{N}(v_i)\cup \{v_i\}}\exp(\text{LeakyReLU}(a^T[Wx_i^{(k-1)}\Vert Wx_t^{(k-1)}]))},\notag
\end{gather} where $W$ is learnable weight matrix, $a$ is the learnable weight vector for attention mechanism, $b$ is the learnable additive bias vector and $\Vert$ is the concatenation operation. Note that GAT follows the same updating scheme as GCN in Eq. (\ref{eq:gcn}), but the normalization coefficients are replaced by learnable attention coefficients.

In a weighted graph, we have \begin{gather}
  x_i^{(k)} = W\cdot\sum_{j=1}^n\hat P_{ij}\alpha_{i,j}  x_j^{(k-1)}+b\\
  \alpha_{i,j} = \frac{\hat P_{ij}\exp(\text{LeakyReLU}(a^T[Wx_i^{(k-1)}\Vert Wx_j^{(k-1)}]))}{\displaystyle\sum_{t=1}^n\hat P_{it} \exp(\text{LeakyReLU}(a^T[Wx_i^{(k-1)}\Vert Wx_t^{(k-1)}]))},\notag
\end{gather} where $\hat P=P+I$ for adding self loops to the GNN.

\subsection{Hyperparameters}\label{sec:app-hp}
The hyperparameter search space is as follows: \begin{itemize}[leftmargin=*]
    \item For non-private MLP, GCN, GraphSAGE and GAT and baseline methods other than \textsc{Solitude}, the learning rate is chosen from $\{10^{-1}, 10^{-2}, 10^{-3}\}$, weight decay is chosen from $\{10^{-3},10^{-4},\allowbreak10^{-5},0\}$, and dropout rate is chosen from $\{10^{-1}, 10^{-2}, 10^{-3}, 0\}$.
    \item For variants of \textsc{Blink}, the learning rate is chosen from $\{10^{-1}, \allowbreak10^{-2}, 10^{-3}\}$, weight decay is chosen from $\{10^{-3},10^{-4},10^{-5},0\}$, dropout rate is chosen from $\{10^{-1}, 10^{-2}, 10^{-3}, 0\}$, and degree privacy parameter $\delta$ is chosen from $\{0.1, 0.3, 0.5, 0.7, 0.9\}$.
    \item For \textsc{Solitude}, both the learning rates for GNN parameters and graph estimation $\hat A$ is chosen from $\{10^{-1}, 10^{-2}, 10^{-3}\}$, weight decay is chosen from $\{10^{-2}, 10^{-3}, 10^{-4}\}$, both $lambda_1$ and $\lambda_2$ are chosen from $\{10^{-2}, 10^{-3}, 10^{-4}, 10^{-5}\}$.
\end{itemize}

After grid search, the selected hyperparameters are the ones with the highest average performance on the validation data over 5 runs, which can be found in \texttt{scripts/output/best\_hp.json} and \texttt{scripts/output/bl\_best\_hp.json} (for baseline methods) of the submitted code base.

\end{document}